\documentclass[11pt]{article}

\usepackage[final]{acl}
\usepackage{times}
\usepackage{latexsym}
\usepackage{pifont}
\usepackage{amsmath, amssymb, amsfonts}
\usepackage{multirow}
\usepackage{tabularx}
\usepackage{enumitem}
\usepackage{xurl}
\usepackage{url}
\usepackage{tcolorbox}
\usepackage{subcaption}
\tcbuselibrary{skins, breakable}
\usepackage[capitalize]{cleveref}
\crefname{appendix}{Appendix}{Appendices}
\Crefname{appendix}{Appendix}{Appendices}
\crefname{table}{Tab.}{Tabs.}    % 小写场景（句中）：单数 Tab. / 复数 Tabs.
\Crefname{table}{Tab.}{Tabs.}    % 句首场景（大写）：单数 Tab. / 复数 Tabs.
\usepackage[table]{xcolor}

\usepackage[T1]{fontenc}
\usepackage[utf8]{inputenc}
\usepackage{booktabs}
\usepackage{microtype}

\usepackage{inconsolata}

\usepackage{graphicx}
\newcommand{\etal}{\emph{et al.}}
\newcommand{\eg}{\emph{e.g.}}
\newcommand{\ie}{\emph{i.e.}}

\title{UR\textsuperscript{2}-MLLM: Uncertainty-aware Revisit Reasoning in Multimodal Large Language Models for Radiology Report Generation}

\author{
\textbf{Yucheng Chen\textsuperscript{1,2,3}},~
\textbf{Yang Yu\textsuperscript{4}},~
\textbf{Jiazhou Zhou\textsuperscript{5}},~
\textbf{Yufei Shi\textsuperscript{1,2}},~
\textbf{Yongying Lan\textsuperscript{1,6}}, \\
\textbf{Yichi Zhang\textsuperscript{1,2,3}},~
\textbf{Liyi Li\textsuperscript{2}},~
\textbf{Si Yong Yeo\textsuperscript{1,2,3}\thanks{Corresponding author.}} \\
\textsuperscript{1}MedVisAI Lab \quad \\
\textsuperscript{2}Lee Kong Chian School of Medicine, Nanyang Technological University, Singapore \\
\textsuperscript{3}Centre of AI in Medicine, Singapore \quad \\ \textsuperscript{4}Institute of Advanced Intelligence and Computing (IAIC), A\textsuperscript{*}STAR, Singapore \quad\\
\textsuperscript{5}AI Thrust, The Hong Kong University of Science and Technology (Guangzhou) \quad\\
\textsuperscript{6}Ruijin Hospital, Shanghai Jiao Tong University School of Medicine, China\\
\texttt{yucheng005@e.ntu.edu.sg} \quad \texttt{siyong.yeo@ntu.edu.sg}  
}

\begin{document}
\maketitle
\begin{abstract}
Radiologists generate diagnostic reports through iterative and selective revisiting of suspicious regions to refine their interpretations. 
Recent multimodal large language models (MLLMs) for radiology report generation (RRG) have shifted from text-only reasoning toward a ``Thinking-with-Images'' paradigm, incorporating visual evidence into the reasoning process. However, existing methods provide static visual evidence without a dynamic revisit mechanism during reasoning, neglecting how radiologists re-examine uncertain observations.
To this end, we propose an Uncertainty-aware Revisit Reasoning MLLM (UR$^{2}$-MLLM) framework that dynamically revisits uncertain regions during reasoning for RRG. UR\textsuperscript{2}-MLLM is first equipped with uncertainty perception by training on an uncertainty-aware dataset. We then construct a multimodal reasoning trajectory dataset together with a detect-and-copy mechanism, which guides when and where to revisit. Finally, a visual grounding reward refines this behavior through reinforcement learning, aligning the revisited regions with corresponding anatomical structures.
Experiments on MIMIC-CXR and IU-Xray show that UR$^{2}$-MLLM achieves state-of-the-art performance, highlighting the value of uncertainty-aware visual revisit reasoning for reliable and clinically aligned report generation.

\begin{figure}
    \centering  \includegraphics[width=0.5\textwidth]{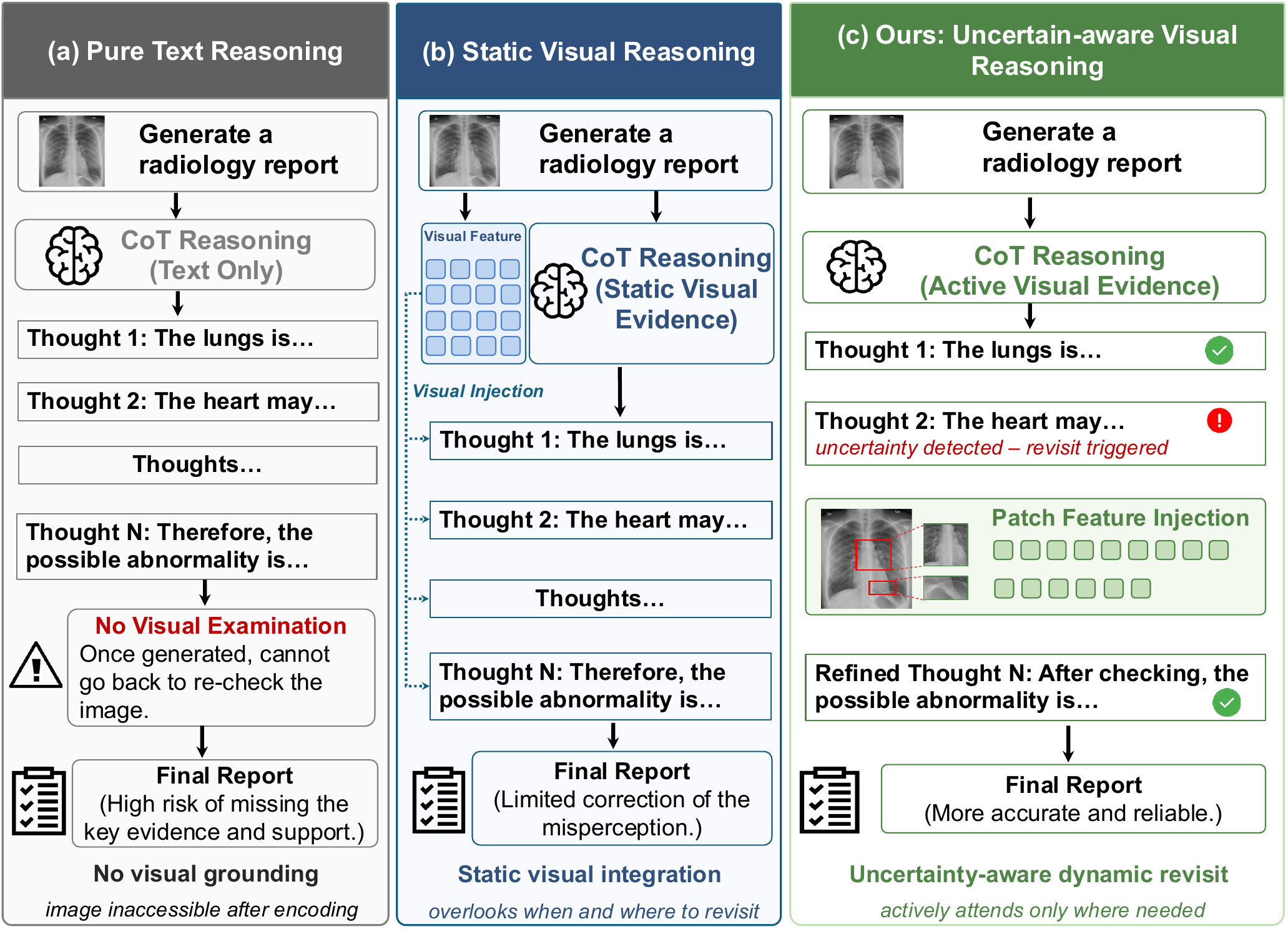} 
    \caption{Three reasoning paradigms for radiology report generation. Our UR\textsuperscript{2}-MLLM triggers uncertainty-aware visual revisits during reasoning.
}   
    \label{fig:coverfig}  
\end{figure}

\end{abstract}

\section{Introduction}

Radiology report generation (RRG) aims to automatically produce diagnostic narratives from medical images~\cite{chen2026riha,chen-emnlp-2020-r2gen}, alleviating the heavy workload of radiologists while improving the quality and consistency of clinical documentation~\cite{DBLP:journals/artmed/IzharIJ25,li2025automatic}. In clinical practice, report generation is inherently a decision-making process under uncertain observations, where radiologists repeatedly revisit suspicious regions, cross-reference visual findings with prior knowledge, and progressively refine their interpretations before committing to a final diagnosis~\cite{van2017visual}. Recent advances in multimodal large language models (MLLMs), together with Chain-of-Thought (CoT)~\cite{DBLP:conf/nips/Wei0SBIXCLZ22} reasoning techniques, have demonstrated strong potential in approximating this iterative reasoning process, enabling more flexible and clinically aligned report generation.

Existing approaches to CoT-based RRG can be broadly categorized into two paradigms. The first follows a text-only reasoning paradigm (\cref{fig:coverfig} (a)), where the model encodes the input image once and conducts the entire reasoning process purely in the textual space~\cite{liu2024medcot,ng2025x,luo2025teaching,fan2025chestx,wang2026curv}. As a result, the reasoning trajectory becomes perceptually decoupled from the image, often leading to error accumulation and hallucinated findings~\cite{liu2025more}. To mitigate the issue, the second paradigm, often referred to as ``Think-with-Images''~\cite{su2025thinkingimagesmultimodalreasoning} (\cref{fig:coverfig}(b)), incorporates static visual evidence throughout the reasoning process. For instance, MAIRA-2~\cite{DBLP:journals/corr/abs-2406-04449} and BoxMed-RL~\cite{jing2025reason} provide the whole image as static visual context and output bbox coordinates as text tokens in the reasoning trace. However, this paradigm approximates the diagnostic process with static visual evidence in a single pass, lacking an explicit mechanism to identify and refine uncertain regions as reasoning progresses. Such an approximation implicitly treats perception as deterministic, discarding the diagnostic uncertainty that is critical in clinically ambiguous cases such as overlapping anatomical structures or subtle pathologies. This often leads to 
over-confident or hallucinated reports, whereas a radiologist would naturally re-examine the relevant regions before finalising the interpretation.
Inspired by this, we ask a fundamental research question: \textbf{\textit{Can an MLLM learn to revisit visual evidence dynamically, deciding when to revisit based on its own uncertainty and where to revisit based on the corresponding regions?}}
 
To bridge this gap, we propose an Uncertainty-aware Revisit Reasoning MLLM~ (UR\textsuperscript{2}-MLLM), a framework that dynamically revisits visual evidence during reasoning, determining \emph{when} and \emph{where} to revisit based on its own uncertainty and the corresponding anatomical regions. 
To this end, UR\textsuperscript{2}-MLLM is implemented through a three-stage pipeline. In Stage~1, we equip the model with uncertainty perception by fine-tuning it on a curated uncertainty-aware dataset, where hedging expressions (\eg,``may'', ``appear'', ``could be'') are aligned with their corresponding anatomical regions on the chest X-ray. This perception capability 
is a prerequisite for the model to later identify \emph{when} a revisit is needed. In Stage~2, we construct a multimodal reasoning trajectory dataset of approximately $2{,}000$ samples that explicitly supervises \emph{when} and \emph{where} to revisit, and introduce a detect-and-copy mechanism that inserts the indices of uncertainty-localized visual patches directly into the reasoning trace. Finally, in Stage~3, we refine this behavior with reinforcement learning using a visual grounding reward, encouraging revisited regions to remain aligned with the corresponding anatomical structures and reducing hallucinated content in the generated reports. Our contributions are as follows.

\begin{itemize}
\item We propose UR\textsuperscript{2}-MLLM, a novel framework to actively revisit visual evidence in an uncertainty-aware manner for RRG, explicitly modelling both when and where to revisit.
\item We introduce a curated uncertainty-aware multimodal reasoning trajectory dataset, along with two core components, a patch-level visual injection mechanism and a visual grounding reward, integrated within a unified three-stage training pipeline.
\item Extensive experiments on IU-Xray and MIMIC-CXR demonstrate that UR\textsuperscript{2}-MLLM achieves state-of-the-art performance, with consistent improvements in both efficiency, faithfulness, and interpretability.

\end{itemize}

\begin{figure*}
    \centering 
    \includegraphics[width=1\textwidth]{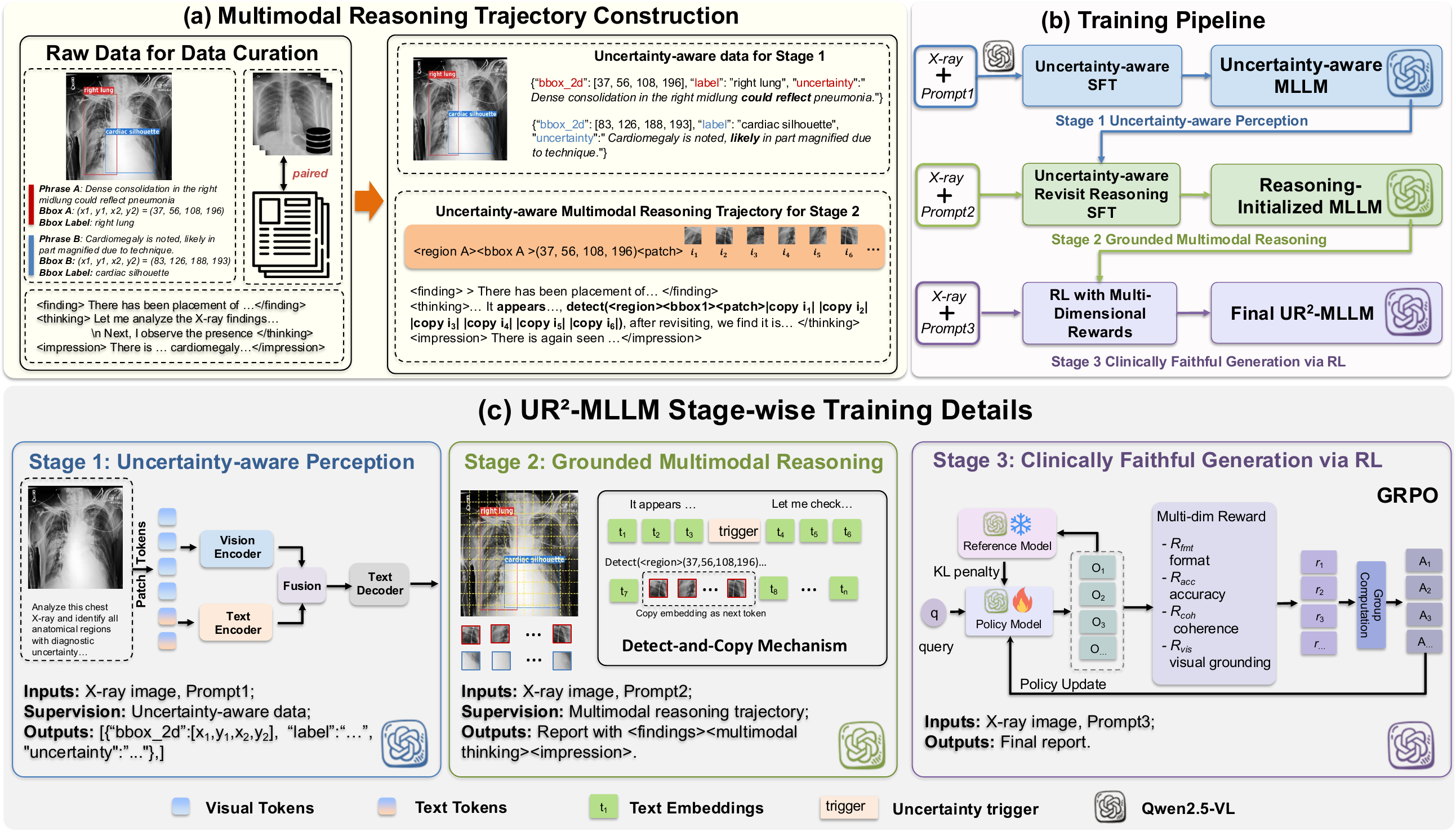} 
   
    \caption{Overall framework of UR\textsuperscript{2}-MLLM, including (a) multimodal reasoning trajectory construction, (b) training pipeline overview, and (c) stage-wise training details.}
    \label{fig2:framework}  
\end{figure*}

\section{Related Work}
\subsection{Uncertainty in RRG}
Modeling diagnostic uncertainty is essential for radiology report generation, since clinical decisions depend on whether a finding is asserted with confidence or expressed with hedging. Existing 
approaches fall into three categories. \textbf{Probabilistic estimation} methods capture uncertainty via Monte Carlo dropout~\cite{DBLP:journals/ijon/WangLXDLTSHZFH24} or variational latent variable models~\cite{najdenkoska2022uncertainty}. \textbf{Consistency-based} methods estimate uncertainty by measuring output variation across perturbed inputs~\cite{zhang2024vl} or sampled generations~\cite{wang2025semantic}. \textbf{Supervision-based} methods extract hedging cues from radiology reports and construct uncertainty-annotated data to train models that explicitly express diagnostic doubt~\cite{wang2026curv, rabaey2025modeling}. Our work falls into the third category, but goes beyond expressing uncertainty in text by using it as a trigger to actively revisit corresponding image regions during reasoning.

\subsection{Reasoning in MLLMs for RRG}
Recent advances in MLLMs have introduced explicit reasoning traces into radiology report generation. Existing approaches fall into two categories based on how visual evidence participates in reasoning. \textbf{Text-based reasoning} methods generate the reasoning trace purely in the textual space after the image has been encoded~\cite{fan2025chestx, luo2025teaching,wang2026curv}, which prevents the model from re-examining the image to verify or refine its claims. \textbf{Static visual reasoning} methods incorporate visual information, either as global image features or as region-level cues such as bounding boxes, alongside the reasoning process~\cite{DBLP:journals/corr/abs-2406-04449, jing2025reason}; however, the visual context remains static once provided throughout generation, offering no mechanism to selectively revisit specific regions when needed. Inspired by how radiologists iteratively re-examine suspicious regions during diagnosis~\cite{kundel1978visual}, we design a framework in which visual revisits are dynamically triggered by 
diagnostic uncertainty, enabling the model to selectively attend to specific image regions throughout the reasoning trace.

\section{Methodology}
In this section, we present UR\textsuperscript{2}-MLLM, a framework for generating radiology reports that are visually grounded and faithful to clinical uncertainty. Emulating the radiologist's diagnostic workflow, our method consists of three training stages, namely \emph{uncertainty-aware perception}, \emph{grounded multimodal reasoning}, and \emph{clinically faithful generation via reinforcement learning (RL)}, supported by a data curation pipeline that produces uncertainty-aware multimodal reasoning trajectories.

\subsection{Task Formulation}

Given a chest X-ray image $X \in \mathbb{R}^{H \times W \times 3}$ and a text instruction $c$ (\ie, a prompt asking the model to generate a structured radiology report), the goal of RRG is to produce a textual report $y = (y_1, y_2, \dots, y_T)$ that 
accurately describes the visual findings and provides a clinical impression. Following standard practice, $y$ typically contains a \emph{Findings} section enumerating observed abnormalities and an \emph{Impression} section summarizing diagnostic conclusions.

Conventional MLLM-based RRG methods learn a MLLM $\Omega_\theta(y \mid X, c)$, where $\theta$ denotes the model parameters, via maximum likelihood 
$\mathcal{L}_{\mathrm{MLE}}(\theta) = -\sum_{t=1}^{T} \log \Omega_\theta(y_t \mid y_{1:t-1}, X, c)$. While effective for fluency, this objective treats $y$ as a flat sequence of tokens and does not explicitly model the radiologist's iterative process of \emph{visual evidence revisit} when uncertainty arises. To address this gap, UR\textsuperscript{2}-MLLM reformulates the objective as maximizing a multi-dimensional reward over uncertainty-aware report generation:
\begin{equation}
\small
    \theta^* = \arg\max_\theta \;\mathbb{E}_{(X, c) \sim \mathcal{D},\, y \sim \Omega_\theta(\cdot \mid X, c)} \bigl[\,R(y, X, c)\,\bigr],
    \label{eq:ur2_objective}
\end{equation}
where $\mathcal{D}$ denotes the training corpus of image–instruction pairs and $R(y, X, c) = \lambda_1 r_{\mathrm{fmt}} + \lambda_2 r_{\mathrm{acc}} + \lambda_3 r_{\mathrm{coh}} + \lambda_4 r_{\mathrm{vis}}$ is a multi-dimensional reward function as detailed in \cref{sec:stage3}.

\subsection{Stage 1: Uncertainty-aware Perception}
\label{sec:stage1}

As shown in \cref{fig2:framework}, Stage 1 equips the base MLLM with the ability to 
localize and articulate diagnostic uncertainty in chest X-ray images. We curate an uncertainty-aware dataset $\mathcal{D}_1$ 
built upon Chest ImaGenome~\cite{PhysioNet-chest-imagenome-1.0.0}, which provides 
bounding-box annotations for anatomical regions in MIMIC-CXR~\cite{PhysioNet-mimic-cxr-2.1.0}. Following CURV~\cite{wang2026curv}, we associate each anatomical region with report phrases containing hedge markers (e.g., \emph{may}, \emph{likely}, \emph{cannot exclude}), yielding a corpus of 
(image, bounding box, label, phrase) tuples. For each image, we aggregate its tuples into a structured target $y$ and pair it with an 
instruction $c$ that prompts the model to enumerate all regions expressing uncertainty. We then finetune the MLLM $\Omega_\theta$ via 
supervised fine-tuning using the autoregressive negative log-likelihood, 
computed only over response tokens:
\begin{equation}
\small
    \mathcal{L}_{\mathrm{S1}}(\theta) = -\,\mathbb{E}_{(X, c, y) \sim \mathcal{D}_1} \sum_{t=1}^{T} \log \Omega_\theta(y_t \mid y_{1:t-1},\, c,\, X).
    \label{eq:stage1_loss}
\end{equation}
The resulting model $\Omega_{\theta_1}$ serves as the initialization for 
Stage 2, having acquired the perceptual prior that diagnostic hedging 
is grounded in specific anatomical regions.

\subsection{Stage 2: Grounded Multimodal Reasoning}
Building on the uncertainty-aware perception acquired in Stage 1, 
Stage 2 trains the model to perform multimodal reasoning that revisits 
image regions whenever diagnostic uncertainty arises. We first describe how we construct grounded reasoning trajectories from the existing report corpus, and then introduce the detect-and-copy mechanism that enables the model to selectively attend to image patches during generation.

\subsubsection{Multimodal Reasoning Trajectory Construction}

We build our multimodal reasoning trajectories upon the text-based 
reasoning traces released by CURV~\cite{wang2026curv}, which provide 
sentence-level CoT annotations for each MIMIC-CXR report in the form of a $\langle\text{findings}\rangle$, 
$\langle\text{thinking}\rangle$, $\langle\text{impression}\rangle$ 
triplet.

To augment these text-only traces with visual revisit events, we scan the $\langle\text{thinking}\rangle$ block sentence by sentence and match each sentence against the phrase–bounding box pairs from Stage 1. When a sentence overlaps with an anatomical phrase, 
we \textbf{(i) }rewrite it to include a hedge cue and a region reference $\langle\text{obj}_n\rangle$ adjacent to the anatomical phrase, 
\textbf{(ii)} insert a structured \texttt{detect} call immediately after, and \textbf{(iii)} append a short revisit summary that resumes the reasoning conditioned on $\langle\text{obj}_n\rangle$. 

To convert the bounding box $b_n$ into a form consumable by the language 
model, we partition each input image into a regular $H' \times W'$ grid of non-overlapping patches and flatten them in row-major order, yielding 
a sequence of $P = H' W'$ patches indexed from $1$ to $P$. The bounding box $b_n$ is then mapped to the set of patch indices $\mathcal{P}_n \subseteq \{1, \dots, P\}$  
that fall within $b_n$. These indices are inserted into the \texttt{detect} call block alongside the anatomical label, enabling the model to reference the revisited region at patch-level granularity. The resulting trajectory is represented as follows:
\begin{equation}
\small
    \underbrace{s_{1:i-1}}_{\text{plain}}
    \to
    \underbrace{\tilde{s}_i}_{\text{hedged}}
    \to
    \underbrace{\mathtt{detect}(g_n, b_n, \mathcal{P}_n)}_{\text{revisit}}
    \to
    \underbrace{r_n}_{\text{summary}}
    \to,
\end{equation}
where $g_n$ and $b_n$ denote the anatomical label and bounding box 
of the revisited region, $\tilde{s}_i$ is the rewritten hedged sentence, and 
$r_n$ is the post-revisit summary, both anchored to the region tag $\langle\text{obj}_n\rangle$. Sentences matching no anatomical phrases remain unchanged. This formulation establishes a 
behavioral pattern in which uncertainty triggers grounded visual revisits, and 
the model learns to continue reasoning after referencing a specific image region. The dataset $\mathcal{D}_2 = \{(X, c, y)\}$ follows the same instruction–response format as Stage 1, with $y$ being 
the full $\langle\text{findings}\rangle$–$\langle\text{thinking}\rangle$–$\langle\text{impression}\rangle$ 
sequence containing region tags, bounding boxes, and patch indices.

\subsubsection{Detect-and-Copy Mechanism}
\begin{figure}
    \centering  \includegraphics[width=0.45\textwidth]{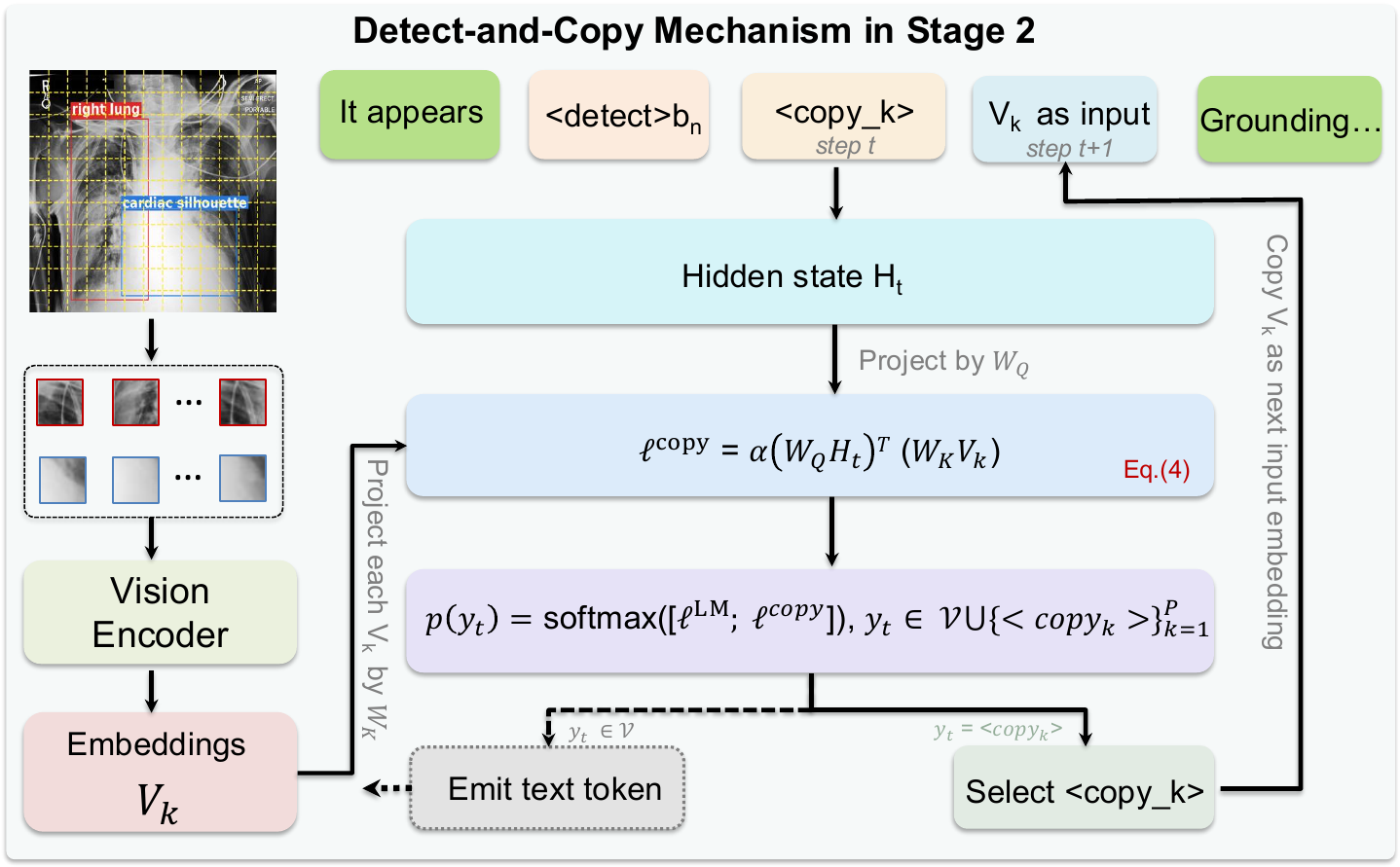} 
    \caption{Details for Detect-and-Copy mechanism. $\mathcal{V}$ denotes the vocabulary and $P$ denotes $\#$patches.}   
    \label{fig:stage2}  
\end{figure}

We introduce a \emph{detect-and-copy} mechanism, inspired by pointer-generator networks~\cite{see2017get,vinyals2015pointer}, which extends the language head to emit image patch indices directly within the reasoning trace as shown in \cref{fig:stage2}.

Let $\mathbf{H}_t \in \mathbb{R}^{d}$ denote the LLM hidden state at decoding step $t$, and $\mathbf{V} \in \mathbb{R}^{P \times d}$ the visual embeddings of the $P$ image patches produced by the vision encoder. We extend the model vocabulary with $P$ copy tokens $\{\langle\text{copy}_k\rangle\}_{k=1}^{P}$, each acting as a learned pointer to one image patch. Linear projections $W_Q$ and $W_K$ map hidden states and visual embeddings into a shared space, and the logit for copy token $k$ at step $t$ is the scaled inner product:
\begin{equation}
\small
    \ell^{\,\text{copy}}_{t, k} = \alpha \cdot \bigl(W_Q \mathbf{H}_t\bigr)^{\!\top} \bigl(W_K \mathbf{V}_k\bigr),
    \label{eq:copy_logit}
\end{equation}
where $\alpha$ is a learnable temperature. These copy logits are concatenated with the standard language-model logits and trained jointly under a single cross-entropy objective, allowing the model to decide at each step whether to emit a text token or a patch pointer. Unlike pointer-generator networks~\cite{see2017get}, no copy gate is required, since the language-model logits and copy logits are defined over disjoint output spaces and normalized by a single softmax. Within a \texttt{detect} call, the model emits the bounding box $b_n$ followed by the copy tokens corresponding to the patches in $\mathcal{P}_n$, producing a learned hard-attention mask anchored to the revisited region. Since the copy logits share their key space with the visual embeddings, subsequent reasoning tokens are naturally grounded in the visual content of the revisited patches. 
Starting from $\Omega_{\theta_1}$, we train Stage 2 with the response-masked autoregressive negative log-likelihood over the extended vocabulary:
\begin{equation}
\small
    \mathcal{L}_{\mathrm{S2}}(\theta) = -\,\mathbb{E}_{(X, c, y) \sim \mathcal{D}_2} \sum_{t=1}^{T} \log \Omega_\theta(y_t \mid y_{1:t-1},\, c,\, X).
    \label{eq:stage2_loss}
\end{equation}
The resulting model $\Omega_{\theta_2}$ serves as the reasoning-initialized MLLM for Stage 3.

\subsection{Stage 3: Clinically Faithful Generation via RL}
\label{sec:stage3}

In this stage, the training data consists of image–report pairs 
$(X, y^*) \in \mathcal{D}_3$ drawn from MIMIC-CXR, where $y^*$ serves 
as the ground-truth reference for reward computation. While Stage 2 
teaches the model to invoke \texttt{detect} calls, it provides no 
signal on whether the revisited region is semantically consistent 
with the reasoning that follows. To close this loop, we introduce a 
\emph{visual grounding reward} $r_{\mathrm{vis}}$ that directly 
supervises the quality of visual revisits. From each trajectory 
$\tau$, we extract the set of revisit events 
$\mathcal{E}(\tau) = \{(b_e, \text{txt}_e)\}_e$, where $b_e$ is the 
bounding box emitted within a \texttt{detect} call and $\text{txt}_e$ 
is the revisit summary sentence emitted immediately after. Two frozen encoders from BiomedCLIP~\cite{zhang2025biomedclipmultimodalbiomedicalfoundation}, $\psi_v$ and 
$\psi_t$, embed the image patch $I_{b_e}$ and the text $\text{txt}_e$ into a shared space. Since cosine similarity between a region and text can be inflated by domain-level visual–textual correlation, we introduce a random-region baseline to calibrate the score. Specifically, we sample $K$ random bounding boxes $\{b_e^{k}\}_{k=1}^{K}$ matching $b_e$ in area and average their similarity scores. The visual-grounding reward $r_{\mathrm{vis}}$ is then defined as:

\begin{equation}
\small
    r_{\mathrm{vis}}(\tau) = \frac{1}{|\mathcal{E}(\tau)|} \sum_{e \in \mathcal{E}(\tau)} \Bigl[\, s(b_e, \text{txt}_e) - \bar{s}(b_e, \text{txt}_e) \,\Bigr],
    \label{eq:vis_reward}
\end{equation}
where $s(b, \text{txt}) = \cos\bigl(\psi_v(I_b), \psi_t(\text{txt})\bigr)$ 
is the cosine similarity between the visual encoding of the image 
patch $I_b$ and the text encoding of $\text{txt}$, and 
$\bar{s}(b, \text{txt}) = \tfrac{1}{K} \sum_{k=1}^{K} s(b^{k}, \text{txt})$ 
is the average similarity over $K$ random bounding boxes 
$\{b^{k}\}_{k=1}^{K}$ of the same area as $b$. We set 
$r_{\mathrm{vis}}(\tau) = 0$ if $\mathcal{E}(\tau) = \emptyset$. The 
subtraction centers the signal around zero, so that the reward 
measures the \emph{specificity} of the predicted bounding box rather than generic domain-level similarity.

We optimize the model with Group Relative Policy Optimization 
(GRPO)~\cite{DBLP:journals/corr/abs-2402-03300}, taking $\Omega_{\theta_2}$ as both 
the initial policy and the frozen reference $\Omega_{\text{ref}}$. The 
visual grounding reward is combined with format adherence 
$r_{\mathrm{fmt}}$, clinical accuracy $r_{\mathrm{acc}}$, and reasoning coherence $r_{\mathrm{coh}}$  into a composite reward $R(\tau) = \sum_k w_k\, r_k(\tau)$, which drives the objective:
\begin{equation}
\small
\begin{aligned}
    \mathcal{L}_{\mathrm{GRPO}}(\theta) = & -\mathbb{E}\Bigl[\tfrac{1}{G}\textstyle\sum_{i=1}^{G} \min(\rho_i A_i,\, \mathrm{clip}_\epsilon(\rho_i) A_i)\Bigr] \\
    & + \beta \, \mathrm{KL}[\Omega_\theta \,\Vert\, \Omega_{\text{ref}}],
\end{aligned}
\label{eq:grpo_loss}
\end{equation}
where $A_i$ is the group-relative advantage of trajectory $\tau_i$ 
and $\rho_i = \Omega_\theta(\tau_i \mid x) / \Omega_{\theta_{\text{old}}}(\tau_i \mid x)$. Here, $\mathrm{clip}_{\epsilon}(\rho_i)=\min(\max(\rho_i,1-\epsilon),1+\epsilon)$ 
denotes clipping $\rho_i$ to $[1-\epsilon,1+\epsilon]$.
The resulting model, denoted UR\textsuperscript{2}-MLLM, generates 
radiology reports that are visually grounded and clinically 
faithful.

\section{Experiments}

\begin{table*}[t]
\centering
\setlength{\tabcolsep}{6pt}
\small
\begin{tabular}{lcccccc}
\toprule
\textbf{Model} & \textbf{BLEU-1} & \textbf{BLEU-2} & \textbf{BLEU-3} & \textbf{BLEU-4} & \textbf{METEOR} & \textbf{ROUGE-L} \\
\midrule
LLaVA-1.5-7B~\cite{liu2024improved}              & 19.09 & 7.46  & 2.81  & 1.25  & 19.16 & 18.36 \\
LLaVA-1.5-7B-SFT-CXR                   & 22.58 & 15.06 & 9.43  & 6.13  & 25.71 & 28.09 \\
HuatuoGPT-Vision-7B~\cite{chen-etal-2024-towards-injecting}   & 19.33 & 9.42  & 4.64  & 1.93  & 26.01 & 20.78 \\
MAIRA-2~\cite{DBLP:journals/corr/abs-2406-04449}                  & 24.94 & 14.12 & 9.01  & 6.14  & 26.78 & 28.65 \\
Qwen2.5-VL-3B~\cite{DBLP:journals/corr/abs-2511-21631}         & 13.09 & 5.42  & 2.08  & 0.89  & 20.81 & 15.23 \\
Gemini 2.5 Pro~\cite{DBLP:journals/corr/abs-2507-06261}                        & 12.54 & 5.20  & 2.25  & 1.05  & 21.19 & 15.01 \\
CURV~\cite{wang2026curv}   & \underline{25.38} & \underline{15.58}  & \underline{9.85}  &  \underline{6.18}  & \textbf{30.43} & \underline{31.19} \\
% ViTAR~\cite{chen2025think} &&&&&&\\
% MedEyes~\cite{zhu2026medeyes} &&&&&&\\
\midrule
\textbf{UR\textsuperscript{2}-MLLM}                          & \textbf{27.79} & \textbf{16.20} & \textbf{10.11} & \textbf{6.73} & \underline{28.66} & \textbf{32.67} \\
\bottomrule
\end{tabular}
\caption{NLG metrics for radiology report generation across different models on MIMIC-CXR(\textbf{Best}, \underline{Second-best}).}
\label{tab:nlg-comparison-mimic}
\end{table*}

\begin{table}[t]
\centering
\setlength{\tabcolsep}{1.5pt}
\footnotesize
\begin{tabular}{l ccc cc}
\toprule
\multirow{2}{*}{\textbf{Model}} & \multicolumn{3}{c}{\textbf{CheXbert}} & \multicolumn{2}{c}{\textbf{RadGraph}} \\
\cmidrule(lr){2-4} \cmidrule(lr){5-6}
& \textbf{Acc.} & \textbf{Ma-F1} & \textbf{Mi-F1} & \textbf{Ent.F1} & \textbf{F1} \\
\midrule
LLaVA-1.5-7B              & 63.25 & 4.94          & 38.54 & 7.61  & 4.95  \\
LLaVA-1.5-7B-SFT-CXR                   & 72.72 & 5.00          & 51.51 & 17.57 & 13.06 \\
HuatuoGPT-Vision-7B   & 71.15 & 5.34          & 48.62 & 15.92 & 9.06  \\
MAIRA-2                  & 67.39 & \underline{6.34} & 46.53 & 25.01 & 17.05 \\
Qwen2.5-VL-3B        & 67.78 & 4.75          & 37.66 & 9.46  & 4.66  \\
Gemini 2.5 Pro                         & 74.35 & 5.35          & 48.45 & 13.09 & 7.71  \\
CURV             & \underline{76.93} & 5.22          & \underline{57.12} & \underline{25.95} & \underline{19.54}  \\

\midrule
\textbf{UR\textsuperscript{2}-MLLM}                          & \textbf{78.67} & \textbf{13.96} & \textbf{62.96} & \textbf{27.72} & \textbf{21.47} \\
\bottomrule
\end{tabular}
\caption{Clinical accuracy metrics on the MIMIC-CXR dataset (\textbf{Best}, \underline{Second-best}).}
\label{tab:mimic_clinical}
\end{table}

\subsection{Experimental Setup}
\noindent\textbf{Dataset.} We curate training data from MIMIC-CXR~\cite{PhysioNet-mimic-cxr-2.1.0} and Chest ImaGenome~\cite{PhysioNet-chest-imagenome-1.0.0}, building an uncertainty-aware perception corpus $\mathcal{D}_1$ for Stage 1 and a multimodal reasoning trajectory corpus $\mathcal{D}_2$ for Stage 2, while Stage 3 reuses MIMIC-CXR dataset without additional annotation. We further test out-of-distribution generalization on the unseen IU X-ray~\cite{demner2016preparing}. More details are provided in \cref{app:dataset}.

\noindent\textbf{Evaluation Metrics.}We evaluate the generated reports from natural language generation (NLG) and clinical efficacy (CE) metrics. CE is assessed by CheXbert~\cite{smit2020combining} (Accuracy, Macro-F1, Micro-F1) and RadGraph~\cite{PhysioNet-radgraph-1.0.0} (Entity F1, complete F1). In \cref{tab:mimic_clinical} and \cref{tab:iu_xray_clinical}, we denote Ma-F1, Mi-F1 and Ent.F1 as Macro F1, Micro F1 and Entity F1, respectively. More descriptions are in \cref{app:evalmetrics}.

\noindent\textbf{Implementation Details.}We initialize all stages from Qwen2.5-VL-3B~\cite{DBLP:journals/corr/abs-2511-21631} and train UR\textsuperscript{2}-MLLM on 4 NVIDIA A6000 GPUs. Stages 1 and 2 use supervised fine-tuning, while Stage 3 applies GRPO. Full training and inference hyperparameters are provided in \cref{app:implementation}.

\begin{table*}[t]
\centering
\small
\begin{tabular}{l cccccc}
\toprule
\textbf{Model} & \textbf{BLEU-1} & \textbf{BLEU-2} & \textbf{BLEU-3} & \textbf{BLEU-4} & \textbf{METEOR} & \textbf{ROUGE-L} \\
\midrule
LLaVA-1.5-7B                & 16.52 & 6.60  & 3.00  & 1.40 & 19.65 & 17.48 \\
LLaVA-1.5-7B-SFT-CXR        & 21.42 & 12.95 & 8.03  & 5.20 & 23.24 & 26.40 \\
HuatuoGPT-Vision-7B         & 19.33 & 10.70 & 6.28  & 2.81 & 31.02 & 23.42 \\
MAIRA-2                     & 26.37 & 15.60 & 9.64  & 6.03 & 25.52 & 31.18 \\
Qwen2.5-VL-3B               & 11.38 & 5.05  & 2.28  & 1.05 & 21.65 & 15.01 \\
CURV  & \underline{29.23} & \underline{18.76}  & \underline{12.08}  & \underline{6.86} & \underline{38.30} & \underline{39.08} \\

\midrule
\textbf{UR\textsuperscript{2}-MLLM}               & \textbf{31.81} & \textbf{19.49} & \textbf{12.94} & \textbf{7.11} & \textbf{40.75} & \textbf{41.20} \\
\bottomrule
\end{tabular}
\caption{NLG metrics for radiology report generation on the IU X-ray dataset (\textbf{Best}, \underline{Second-best}).}
\label{tab:iu_xray_results}
\end{table*}

\begin{table}[t]
\centering
\setlength{\tabcolsep}{3.5pt}
\footnotesize
\begin{tabular}{l ccc cc}
\toprule
\multirow{2}{*}{\textbf{Model}} & \multicolumn{3}{c}{\textbf{CheXbert}} & \multicolumn{2}{c}{\textbf{RadGraph}} \\
\cmidrule(lr){2-4} \cmidrule(lr){5-6}
& \textbf{Acc.} & \textbf{Ma-F1} & \textbf{Mi-F1} & \textbf{Ent.F1} & \textbf{F1} \\
\midrule
LLaVA-1.5-7B           & 72.03 & 4.66          & 46.81 & 13.37 & 8.76  \\
LLaVA-1.5-SFT          & 76.34 & 5.84          & 53.33 & 16.19 & 10.31 \\
HuatuoGPT-V-7B         & 89.89 & 5.57          & 67.07 & 22.98 & 13.96 \\
MAIRA-2                & 88.74 & \underline{6.22} & 70.75 & 34.53 & 24.01 \\
Qwen2.5-VL-3B          & 80.64 & 4.92          & 49.47 & 12.70 & 6.26  \\
CURV & \underline{91.56} & 5.86          & \textbf{74.36} & \underline{36.99} & \underline{25.65} \\
\midrule
\textbf{UR\textsuperscript{2}-MLLM}          & \textbf{92.58} & \textbf{8.45} & \underline{74.15} & \textbf{37.87} & \textbf{25.95} \\
\bottomrule
\end{tabular}
\caption{Clinical accuracy metrics on the IU X-ray dataset (\textbf{Best}, \underline{Second-best}).}
\label{tab:iu_xray_clinical}
\end{table}

\subsection{Comparison Results}

We evaluate UR\textsuperscript{2}-MLLM on the MIMIC-CXR~\cite{PhysioNet-mimic-cxr-2.1.0} test split against a diverse set of baselines, including general-purpose MLLMs (LLaVA-1.5-7B~\cite{liu2024improved}, Qwen2.5-VL-3B~\cite{DBLP:journals/corr/abs-2511-21631}, Gemini 2.5 Pro~\cite{DBLP:journals/corr/abs-2507-06261}), medical-domain MLLMs (HuatuoGPT-Vision-7B~\cite{chen-etal-2024-towards-injecting}, LLaVA-1.5-7B-SFT-CXR), the grounded RRG model MAIRA-2~\cite{DBLP:journals/corr/abs-2406-04449} and CURV~\cite{wang2026curv} as the closest reasoning-oriented baseline. As shown in \cref{tab:nlg-comparison-mimic}, UR\textsuperscript{2}-MLLM achieves the best performance on five out of six NLG metrics, with BLEU-1, BLEU-2, BLEU-3, BLEU-4, and ROUGE-L all surpassing the strongest baseline CURV. On METEOR, our model scores 28.66, slightly below CURV's 30.43 but remaining competitive against all other MLLM baselines. On the clinical efficacy metrics in \cref{tab:mimic_clinical}, UR\textsuperscript{2}-MLLM achieves 78.67 CheXbert Acc., 13.96 Ma-F1, 62.96 Mi-F1, 27.72 RadGraph Ent.F1, and 21.47 RadGraph F1, outperforming CURV across all 5 metrics. The improvement is particularly notable on the clinically-oriented metrics such as RadGraph F1, suggesting that grounding the reasoning trace in visual evidence through adaptive revisit produces reports that are more clinically faithful, not merely more fluent.

\begin{figure}
    \centering  \includegraphics[width=0.45\textwidth]{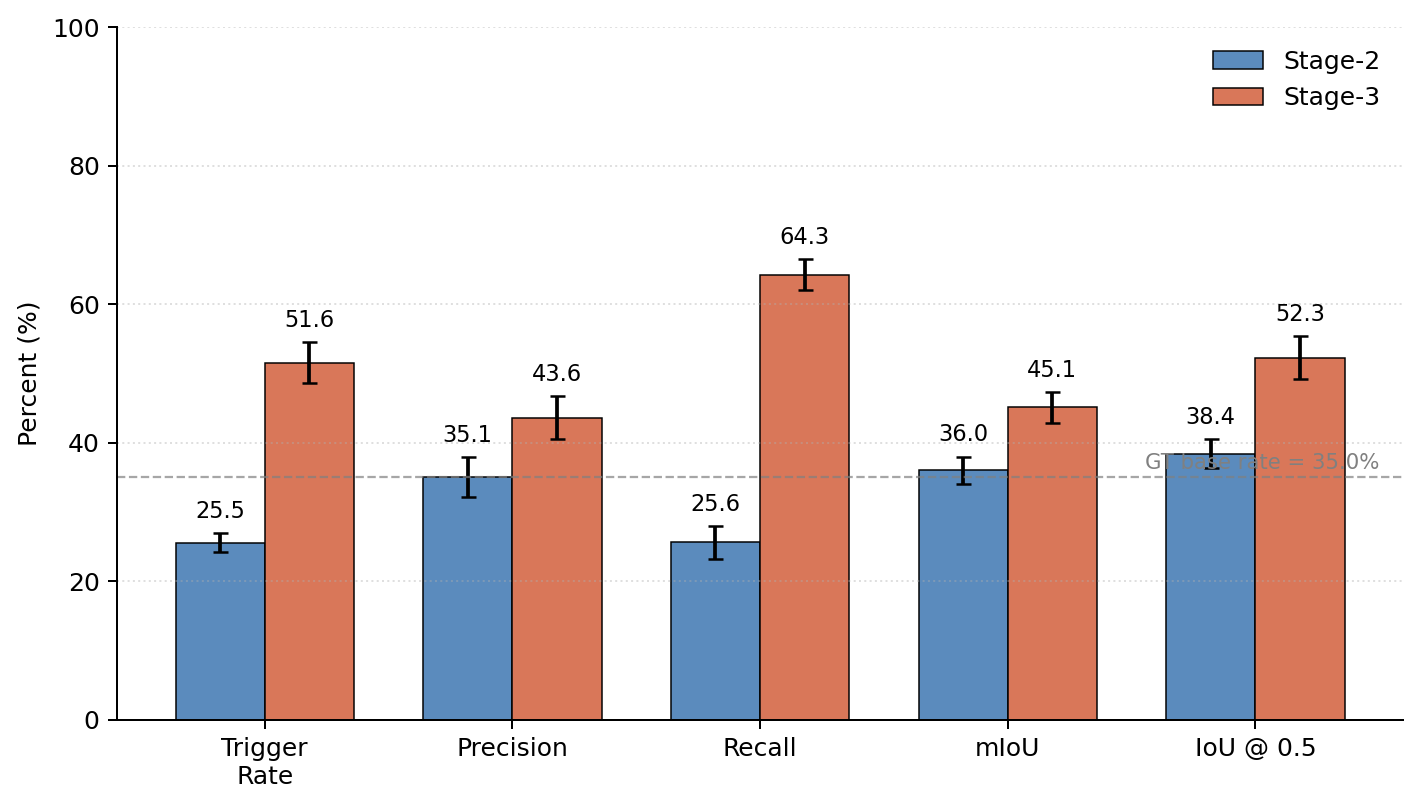} 
    \caption{Revisit-quality on the MIMIC-CXR.}   
    \label{fig:reasoning-evaluation}  
\end{figure}

\subsection{Out-of-Distribution Evaluation}

To assess generalization beyond the training distribution, we evaluate UR\textsuperscript{2}-MLLM on IU X-ray~\cite{demner2016preparing}, which is unseen during all three training stages and differs from MIMIC-CXR in patient population, acquisition protocol, and reporting style. As shown in \cref{tab:iu_xray_results}, UR\textsuperscript{2}-MLLM achieves the best performance across all six NLG metrics, reaching 31.81 BLEU-1, 7.11 BLEU-4, 40.75 METEOR, and 41.20 ROUGE-L, consistently outperforming CURV and all other MLLM baselines. On the clinical efficacy side (\cref{tab:iu_xray_clinical}), UR\textsuperscript{2}-MLLM attains 92.58 CheXbert Acc. and 8.45 Ma-F1, both the highest among all baselines, along with the top RadGraph Ent.F1 (37.87) and RadGraph F1 (25.95). The only metric on which our model does not take the lead is CheXbert Mi-F1 (74.15 vs.\ CURV's 74.36), with the gap being marginal. These results demonstrate that the gains brought by adaptive visual revisit are not specific to the MIMIC-CXR distribution, but transfer to an unseen dataset, indicating that the model has learned a transferable strategy of grounding reasoning in visual evidence rather than overfitting to dataset-specific patterns.

\subsection{Ablation Study}

\begin{table*}[t]
\centering
\setlength{\tabcolsep}{5pt}
\small
\begin{tabular}{l cccccc ccc}
\toprule
\multirow{2}{*}{\textbf{Model}} & \multicolumn{6}{c}{\textbf{NLG}} & \multicolumn{3}{c}{\textbf{CE}} \\
\cmidrule(lr){2-7} \cmidrule(lr){8-10}
& \textbf{BL-1} & \textbf{BL-2} & \textbf{BL-3} & \textbf{BL-4} & \textbf{MTR} & \textbf{RG-L} & \textbf{F1} & \textbf{Rec.} & \textbf{Prec.} \\
\midrule
Base (Qwen2.5-VL-3B)                                       & 13.09 & 5.42 & 2.08 & 0.89 &20.81  & 15.23 & 4.75 & - & - \\
+ S1                                       & 9.88 &4.31  &1.91  &1.02  &20.33  &10.74  &12.35  &28.06  &13.47  \\
+ S2                                       & 23.66 &13.88  &8.91  & 5.96 & 21.77 &18.69  & 11.66 & 24.50 &14.68  \\
+ S1 + S2 (w/o visual revisit)             &10.72  &5.22  &2.43  &1.10  &18.57  &14.59  &3.87  & - & - \\
+ S1 + S2 (w/ static vision)             &16.13  &5.01  &5.19  &1.98  &17.10  &14.97  &6.13  &21.30  &14.96  \\
+ S1 + S2                                  & \underline{26.98} &\underline{15.61}  &\underline{9.75}  &\underline{6.44}  &23.36  &\underline{21.36}  &\underline{13.55}  &\underline{32.34}  &17.25  \\
+ S1 + S2 + S3 (w/o $R_\text{rev}$)        &25.26  &13.94  & 8.03 &5.07  &\underline{24.96}  & 16.11 &  13.00& 30.64 & \textbf{\underline{26.09}} \\
+ S1 + S2 + S3 (\textbf{UR\textsuperscript{2}-MLLM})         & \textbf{27.79} &\textbf{16.20}  &\textbf{10.11}  &\textbf{6.73}  &\textbf{28.66}  &\textbf{32.67}&\textbf{13.96}  &\textbf{32.39}  &\underline{18.03}  \\
\bottomrule
\end{tabular}
\caption{Ablation study of UR\textsuperscript{2} on MIMIC-CXR (\textbf{Best}, \underline{Second-best}). S1, S2, S3 denotes Stage1, Stage2, and Stage3, respectively. F1 here denotes CheXbert Macro F1.}
\label{tab:ablation}
\end{table*}

\begin{figure*}
    \centering
    \includegraphics[width=0.9\textwidth, height=0.2\textheight]{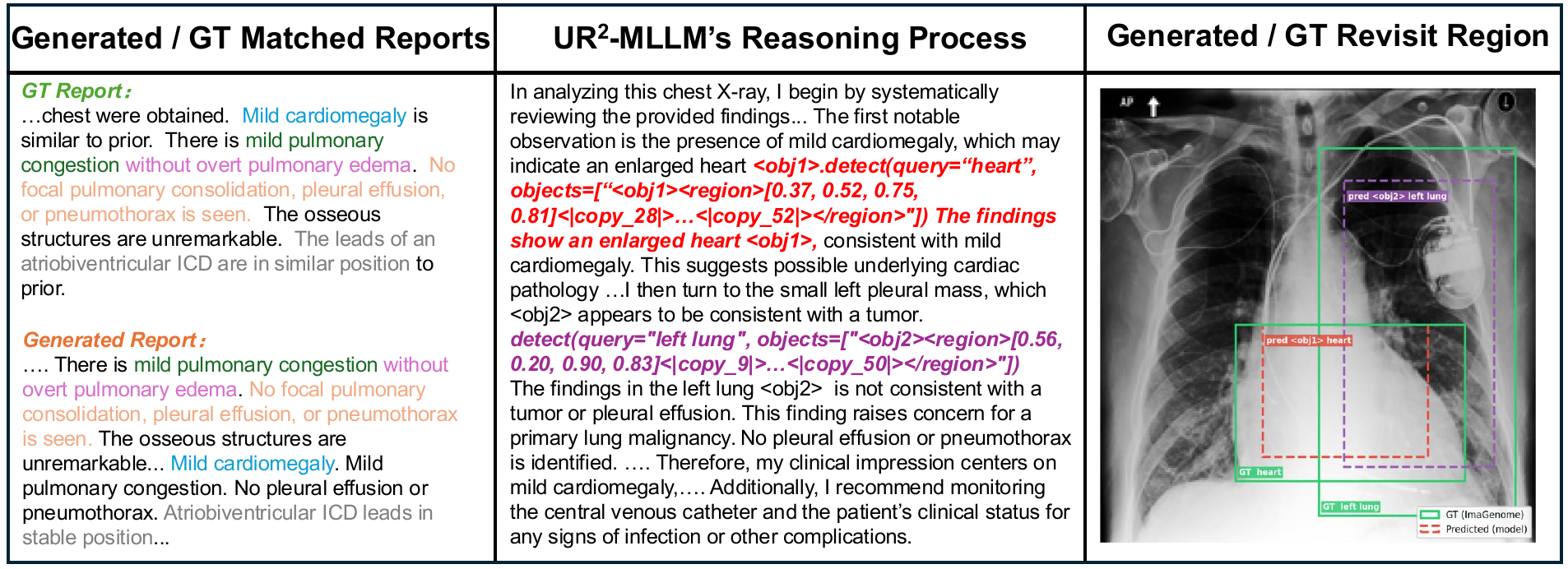}
    \caption{A success case of UR\textsuperscript{2}-MLLM. Left: ground-truth and generated reports with matched findings highlighted. Middle: reasoning trace where uncertainty triggers \texttt{detect} calls. Right: predicted bounding boxes (red, purple) versus ground-truth (green) from Chest ImaGenome.}
    \label{fig:vis}
\end{figure*}

To validate each design choice in UR\textsuperscript{2}-MLLM, we conduct an ablation study on MIMIC-CXR, examining the contribution of each training stage, dynamic visual revisit versus static visual injection, and the effect of the visual grounding reward $r_{\mathrm{vis}}$, as reported in \cref{tab:ablation}. Starting from the Qwen2.5-VL backbone (Base), Stage 1 alone (+S1) slightly degrades NLG metrics, since this stage focuses on structured uncertainty enumeration rather than full report generation, but substantially improves clinical F1 (4.75 $\to$ 12.35). Adding Stage 2 (+S1+S2) yields a large NLG gain as the model produces complete reasoning trajectories, and Stage 3 further pushes performance to 27.79 BLEU-1, 28.66 METEOR, and 13.96 clinical F1. To isolate the value of dynamic revisit, we further consider two Stage 2 variants: w/o visual revisit, which disables the detect-and-copy mechanism, and w/ static vision, where the full image together with the target bounding box is provided as a fixed visual context that the reasoning attends to via cross-attention throughout, without dynamically revisiting specific regions. The w/o visual revisit variant drops sharply on both NLG (BLEU-1: 10.72) and clinical F1 (3.87), and the w/ static vision variant reaches only 16.13 BLEU-1 and 6.13 clinical F1, still far behind our dynamic revisit design. Finally, removing $r_{\mathrm{vis}}$ from Stage 3 degrades both NLG and clinical efficacy (BLEU-1: 2.53$\downarrow$, clinical F1: 0.96 $\downarrow$, Recall: 1.75$\downarrow$), confirming that supervising \emph{where} the model looks is not sufficient on its own and that the reward plays a complementary role in aligning the revisited region with the textual claim. Overall, these results validate that each stage and design component is essential, and that dynamic visual revisit anchored by uncertainty and reinforced by a visual grounding reward is the key ingredient behind UR\textsuperscript{2}-MLLM's performance.

\subsection{Visual Reasoning Evaluation}
To assess whether UR\textsuperscript{2}-MLLM genuinely learns when and where to revisit visual evidence rather than imitating the surface form of reasoning traces, we evaluate its revisit behavior on the MIMIC-CXR test split (n=3,858). We report the trigger rate, the precision and recall of the triggering decision, and the spatial accuracy of the revisited region measured by mIoU and IoU@0.5. The ground-truth base rate of samples genuinely requiring a revisit is 35.0\%. As shown in \cref{fig:reasoning-evaluation}, Stage 3 consistently improves over Stage 2 across all five metrics. The trigger rate rises from 25.5\% to 51.6\%, crossing the base rate and shifting the model from under-triggering to actively invoking revisits. The recall improves substantially from 25.6\% to 64.3\%. These indicate that the reinforcement learning stage instills a visually grounded revisit policy rather than a purely linguistic one. While these improvements are clear, the absolute precision and spatial overlap remain limited.

\subsection{Case Study}
\cref{fig:vis} presents a representative success case from MIMIC-CXR. The report generated by UR\textsuperscript{2}-MLLM closely aligns with the ground truth across all key findings, including mild cardiomegaly, mild pulmonary congestion, the absence of pleural effusion or pneumothorax, and the atriobiventricular ICD. The middle column further reveals the model's reasoning process: when diagnostic uncertainty arises, the model triggers \texttt{detect} calls to revisit specific anatomical regions, first the heart ($\langle\text{obj}_1\rangle$) and then the left lung ($\langle\text{obj}_2\rangle$). As shown in the right column, the predicted bounding boxes (red and purple) are spatially well-aligned with the Chest ImaGenome ground-truth boxes (green), confirming that the model attends to the correct anatomical structures during revisit. Additional cases are provided in \cref{app:casestudy}.

\section{Conclusion}
\label{sec:conclusion}

We present UR\textsuperscript{2}-MLLM, an uncertainty-aware multimodal large language model for radiology report generation that emulates how radiologists iteratively revisit suspicious regions during diagnosis. Unlike prior methods that rely on static visual evidence, UR\textsuperscript{2}-MLLM treats diagnostic uncertainty as an explicit trigger for visual revisit, deciding \emph{when} to look again and \emph{where} to focus as reasoning unfolds. 
The model progressively learns to perceive uncertainty, ground reasoning in localized visual evidence, and refine revisit behavior via reinforcement learning. Across in-distribution and out-of-distribution benchmarks, UR\textsuperscript{2}-MLLM achieves consistent improvements on both linguistic and clinical efficacy metrics, with further analysis showing that these gains stem from genuinely grounded visual revisit rather than imitation of surface reasoning patterns.
We hope this work encourages further research on visually grounded, uncertainty-aware reasoning for trustworthy medical report generation.

\section*{Limitations}
\label{sec:limitations}

While UR\textsuperscript{2}-MLLM demonstrates the value of uncertainty-triggered visual revisit, several limitations remain. First, our notion of diagnostic uncertainty is derived from textual hedging cues, which provides only a coarse and weakly constrained signal. A more clinically grounded formulation of uncertainty, potentially informed by radiologist confidence ratings or calibrated 
probability estimates, could yield a stronger and more reliable trigger. Second, the multimodal reasoning trajectory corpus used in Stage 2 is relatively small ($\sim$2K trajectories) and is partly 
derived from LLM-rewritten reasoning traces, which may introduce upstream biases. Scaling up the dataset with higher-quality and expert-validated trajectories is an important direction for future work. Third, the spatial grounding in our pipeline relies on the 
silver-standard anatomical annotations of Chest ImaGenome, which constrains the precision of the revisit supervision and limits applicability to settings where such region-level annotations are unavailable. Finally, our evaluation is confined to chest X-ray benchmarks with automatic metrics. The absence of external clinical validation and human-in-the-loop assessment by radiologists leaves 
the real-world clinical reliability of the generated reports an open question. We view UR\textsuperscript{2}-MLLM as a meaningful but preliminary step toward uncertainty-aware visual reasoning, and 
addressing these limitations is essential for clinical deployment.

\section*{Ethics Considerations}
This research complies with established ethical standards. All datasets employed are publicly
available and utilized solely for their intended research purposes. These datasets contain no personally identifiable information or sensitive content, thereby posing no risks to privacy or confidentiality. In addition, the study did not involve human subjects or annotators.
% \section*{Acknowledgments}

\bibliography{custom}

\clearpage
\appendix

\begin{center}
{\Large\bfseries Appendix}
\end{center}

\section{Experimental Setup}
\label{app:exp}
\subsection{Dataset}
\label{app:dataset}
We curate our training data from two publicly available resources. MIMIC-CXR~\cite{PhysioNet-mimic-cxr-2.1.0} is a large-scale chest X-ray dataset containing over 370,000 images paired with free-text radiology reports. Chest ImaGenome~\cite{PhysioNet-chest-imagenome-1.0.0} provides anatomical bounding box annotations and structured scene graphs for MIMIC-CXR studies. Built upon these two resources, we construct two stage-specific corpora: an uncertainty-aware perception corpus $\mathcal{D}_1$ containing approximately 101k samples used in Stage 1, and a multimodal reasoning trajectory corpus $\mathcal{D}_2$ containing around 2,000 curated trajectories used in Stage 2. Stage 3 reinforcement learning samples prompts from MIMIC-CXR without requiring additional annotation. All experiments adopt the train/test split defined by PromptMRG~\cite{jin2024promptmrg}, with 3,858 studies held out for evaluation. To further assess out-of-distribution generalization, we additionally evaluate on IU X-ray~\cite{demner2016preparing}, a smaller chest X-ray dataset that the model has not been exposed to during training. The statistic details for our curation datasets can be found in \cref{tab:supp-dataset-sizes}.

\begin{table*}[h]
\centering
\small
\begin{tabular}{llrr}
\toprule
\textbf{Dataset} & \textbf{Split} & \textbf{Image-level} & \textbf{Item-level} \\
\midrule
\multicolumn{4}{l}{\textit{Stage~1 SFT data}} \\
\quad uncertainty-aware dataset       & full                  & 101{,}457 images & 542{,}834 bbox items \\
\quad\quad ($\sim$5.35 bbox / image, median 4, max 39)  & --- & ---                 & ---                    \\
\addlinespace
\multicolumn{4}{l}{\textit{Stage~2 SFT data}} \\
\quad multimodal\_reasoning\_trajectory  & full                  & 1{,}909 images & 3{,}172 revisit anchors \\
\quad\quad Type A (with detect() calls) & ---                 &      1{,}292 images & 3{,}172 anchors          \\
\quad\quad Type B (no detect() calls)   & ---                 &        617 images   & 0 anchors                  \\
\quad train / val (5\% holdout)      & train / val           & 1{,}814 / 95   & ---                    \\
\addlinespace
\multicolumn{4}{l}{\textit{Stage~3 RL data}} \\
\quad MIMIC-CXR (\cite{PhysioNet-mimic-cxr-2.1.0})  & train  & 227{,}835 samples       & --- \\
\quad                                & test                  &    3{,}858 samples     & --- \\
\addlinespace
\multicolumn{4}{l}{\textit{Out-of-domain evaluation}} \\
\quad IU-Xray (\cite{demner2016preparing}) & train                & 2{,}069 samples    & ---       \\
\quad                                & val                   &    296 samples     &    ---      \\
\quad                                & test                  &    590 samples     & ---       \\
\bottomrule
\end{tabular}
\caption{Dataset sizes used by UR\textsuperscript{2}-MLLM.  ``Image-level'' counts the unique image-report pairs that become training (or evaluation) samples; ``item-level'' counts the finer-grained annotations derived from each pair, since a single MIMIC-CXR study yields multiple bbox-uncertainty annotations during Stage 1 curation and a single ImaGenome-anchored sentence yields multiple revisit regions during Stage 2 curation, so item-level counts are strictly larger than image-level counts wherever applicable.  Stage 3 uses MIMIC-CXR image-report pairs without additional annotation. IU-Xray is used only for out-of-domain evaluation; its ``item-level'' column counts the individual frontal and lateral DICOM images, since a single report covers both views.}
\label{tab:supp-dataset-sizes}
\end{table*}

\subsection{Evaluation Metrics}
\label{app:evalmetrics}
We evaluate the generated reports along two complementary axes. 
Natural language generation (NLG) quality is measured against 
the ground-truth report using BLEU-1/2/3/4~\cite{papineni2002bleu}, 
METEOR~\cite{banerjee2005meteor}, and ROUGE-L~\cite{lin-2004-rouge}. Clinical efficacy (CE) evaluates the factual agreement on the 14 CheXpert pathologies, with predicted labels extracted by the CheXbert labeler~\cite{smit2020combining}. We report Accuracy, Macro-F1, 
and Micro-F1. We additionally report Entity F1 and complete F1 from RadGraph~\cite{PhysioNet-radgraph-1.0.0} to assess clinical entity alignment.

\subsection{Implementation Details}
\label{app:implementation}
We initialize all stages from Qwen2.5-VL-3B~\cite{DBLP:journals/corr/abs-2511-21631} and train UR\textsuperscript{2}-MLLM on 4 NVIDIA A6000 GPUs with BF16 mixed precision. We use AdamW with a cosine learning rate schedule and 5\% warm-up. Stages 1 and 2 are trained via supervised fine-tuning for 3 epochs each, with a learning rate of $1\!\times\!10^{-5}$ and an effective batch size of 32. Stage 3 further refines 
the model via GRPO for 1 epoch over a sampled subset of MIMIC-CXR prompts, with $G = 4$ trajectories per prompt, a KL coefficient $\beta = 0.04$, and clip range $\epsilon = 0.2$. At inference, we use nucleus sampling with temperature $T = 0.7$ and top-$p = 0.9$. Our code will be released upon acceptance.

\section{Prompt details}
\subsection{Data Curation Prompts}
\label{sec:supp-curation-prompts}

This section illustrates every LLM prompt in details used to construct the CURV Stage~1 and Stage~2 datasets.  All prompts target a locally hosted \texttt{Qwen/Qwen2.5-7B-Instruct} model exposed as an OpenAI-compatible endpoint, with $\text{temperature}\in[0.0, 0.4]$, JSON-mode responses,
and $\leq3$ retries on parse failure.  Stage~1 prompts (\S\ref{ssec:s1-curation}) filter MIMIC-CXR reports and label
uncertainty at the sentence and phrase level; Stage~2 prompts (\S\ref{ssec:s2-curation}) rewrite uncertainty sentences into hedged form and emit visual-revisit summaries.  A summary of the pipeline is given in \cref{tab:supp-curation-summary}.

% =====================================================================
\subsubsection{Stage~1 dataset curation prompts}
\label{ssec:s1-curation}

The Stage~1 pipeline turns raw MIMIC-CXR reports into the uncertainty-aware perception dataset $\mathcal{D}_1$ used for Stage~1 SFT.
Three LLM call types are used: a report section parser (three header-conditional variants, \cref{tab:supp-prompt-s1a}),
a section-level uncertainty extractor (\cref{tab:supp-prompt-s1b}),
and a per-phrase uncertainty classifier (\cref{tab:supp-prompt-s1c}).

% ---------------------------------------------------------------------
\begin{table*}[!htbp]
\centering
\small
\renewcommand{\arraystretch}{1.25}
\begin{tabular}{@{} >{\centering\arraybackslash}p{0.13\textwidth} p{0.83\textwidth} @{}}
\toprule
\multicolumn{2}{>{\centering\arraybackslash}p{0.96\textwidth}}{\textit{(a) Variant: impression label only --- look for unlabelled findings content}} \\
\midrule
\textbf{system content} &
You are a radiology report parser.
\par\smallskip
The report below has a labeled \texttt{IMPRESSION} section but \emph{no}
labeled \texttt{FINDINGS} section.  Determine whether the report also
contains radiographic observations (findings content)---any text that
describes the appearance of anatomical structures (e.g.\ lungs, heart,
bones, pleura) using imaging language.
\par\smallskip
Answer \emph{YES} (true) if the report contains any of the following
outside the IMPRESSION section:\newline
\hspace*{1em}--- one or more sentences describing the appearance of
anatomical structures;\newline
\hspace*{1em}--- observations using terms like ``clear'', ``normal'',
``enlarged'', ``opacity'', ``effusion'', ``no evidence of'', ``stable'',
``unchanged'', etc.;\newline
\hspace*{1em}--- a descriptive paragraph providing detail beyond the
impression summary.
\par\smallskip
Answer \emph{NO} (false) only if the report contains nothing but the
IMPRESSION section and possibly a brief technique / clinical-history
note with no imaging observations whatsoever.
\par\smallskip
Reply only with JSON: \texttt{\{"has\_findings": true\}} or
\texttt{\{"has\_findings": false\}}.  No explanation.
\\
\midrule
\multicolumn{2}{>{\centering\arraybackslash}p{0.96\textwidth}}{\textit{(b) Variant: findings label only --- look for unlabelled impression / conclusion}} \\
\midrule
\textbf{system content} &
You are a radiology report parser.
\par\smallskip
The report below has a labeled \texttt{FINDINGS} section but \emph{no}
labeled \texttt{IMPRESSION} section.  Determine whether the report also
contains an impression or conclusion---any text that provides a
clinical interpretation, summary, or diagnosis based on the findings.
\par\smallskip
Answer \emph{YES} (true) if the report contains any of the
following:\newline
\hspace*{1em}--- one or more sentences summarising the overall clinical
interpretation;\newline
\hspace*{1em}--- a concluding statement giving a diagnosis or
differential;\newline
\hspace*{1em}--- a final paragraph or sentence that reads as a summary /
conclusion rather than a detailed observation.
\par\smallskip
Answer \emph{NO} (false) only if the report ends with detailed findings
observations and contains no interpretive summary or conclusion
anywhere.
\par\smallskip
Reply only with JSON: \texttt{\{"has\_impression": true\}} or
\texttt{\{"has\_impression": false\}}.  No explanation.
\\
\midrule
\multicolumn{2}{>{\centering\arraybackslash}p{0.96\textwidth}}{\textit{(c) Variant: no header labels --- look for findings and impression jointly}} \\
\midrule
\textbf{system content} &
You are a radiology report parser.
\par\smallskip
The report below has \emph{no} labeled FINDINGS or IMPRESSION sections.
Determine whether the report contains \emph{both} types of content,
even without labels:
\par\smallskip
\hspace*{1em}A) findings content: text describing the radiographic
appearance of anatomical structures (lungs, heart, bones, \ldots);\newline
\hspace*{1em}B) impression / conclusion content: text that summarises
the overall clinical interpretation, diagnosis, or conclusion.
\par\smallskip
Answer \emph{YES} (true) if you can identify both types of content in
the report, even if they are not in separate paragraphs.
\par\smallskip
Answer \emph{NO} (false) if the report contains only one type of
content, or if it is too brief to clearly contain both.
\par\smallskip
Reply only with JSON: \texttt{\{"has\_both": true\}} or
\texttt{\{"has\_both": false\}}.  No explanation.
\\
\midrule
\multicolumn{2}{>{\centering\arraybackslash}p{0.96\textwidth}}{\textit{User content (shared across all three variants)}} \\
\midrule
\textbf{user content} &
\texttt{Report:}\newline
\texttt{\{report\}}
\\
\bottomrule
\end{tabular}
\caption{Stage 1A prompt --- report section parser (three
header-conditional variants).  All three variants share the same user
content, shown last; \texttt{\{report\}} is replaced with the report
body, truncated to 4000 characters.  The system content differs by
which section headers are present in the raw report.}
\label{tab:supp-prompt-s1a}
\end{table*}

% ---------------------------------------------------------------------
\begin{table*}[!htbp]
\centering
\small
\renewcommand{\arraystretch}{1.25}
\begin{tabular}{@{} >{\centering\arraybackslash}p{0.13\textwidth} p{0.83\textwidth} @{}}
\toprule
\textbf{system content} &
You are a medical text analysis expert specialising in uncertainty
detection.  Analyse the provided radiology report section and identify
all uncertainty expressions---words or phrases that indicate diagnostic
uncertainty, probability, or hedging (e.g.\ likely, possibly, may,
consistent with, cannot exclude, suspicious for).
\\
\midrule
\textbf{user content} &
Analyse the following radiology report section and identify uncertainty
expressions.
\par\smallskip
\texttt{Text: ``\{text\}''}
\par\smallskip
Respond with a JSON object containing exactly these
keys:\newline
\hspace*{1em}\texttt{"uncertainty\_phrases"}: list of uncertainty
expressions found (strings);\newline
\hspace*{1em}\texttt{"confidence\_level"}: \texttt{"high"},
\texttt{"medium"}, or \texttt{"low"};\newline
\hspace*{1em}\texttt{"uncertainty\_type"}: \texttt{"structural"},
\texttt{"semantic"}, or \texttt{"both"};\newline
\hspace*{1em}\texttt{"reasoning"}: one-sentence explanation.
\par\smallskip
If no uncertainty expressions are found, return\newline
\hspace*{1em}\texttt{\{"uncertainty\_phrases": [], "confidence\_level":
"low", "uncertainty\_type": "structural", "reasoning": "No uncertainty
found."\}}
\\
\bottomrule
\end{tabular}
\caption{Stage 1B prompt --- section-level uncertainty extraction.
The findings text is truncated to 2000 characters before substitution
into \texttt{\{text\}}.}
\label{tab:supp-prompt-s1b}
\end{table*}

% ---------------------------------------------------------------------
\begin{table*}[!htbp]
\centering
\small
\renewcommand{\arraystretch}{1.25}
\begin{tabular}{@{} >{\centering\arraybackslash}p{0.13\textwidth} p{0.83\textwidth} @{}}
\toprule
\textbf{system content} &
You are a radiology NLP expert.  Decide whether a radiology report
sentence expresses diagnostic uncertainty about a finding.
\par\smallskip
Uncertainty markers include: \emph{likely, may, might, could, possibly,
probably, suspicious, suggestive of, cannot exclude, consistent with,
versus, questionable, equivocal, appears, consider, reflect, represent},
etc.
\par\smallskip
Reply only with a JSON object: \texttt{\{"is\_uncertain": true\}} or
\texttt{\{"is\_uncertain": false\}}.  No explanation.
\\
\midrule
\textbf{user content} &
\texttt{Sentence: ``\{phrase\}''}
\\
\bottomrule
\end{tabular}
\caption{Stage 1C prompt --- per-phrase uncertainty classifier
(binary).  \texttt{\{phrase\}} is an ImaGenome scene-graph phrase
(a noun-phrase or short clause) anchored to a single anatomical
bounding box.}
\label{tab:supp-prompt-s1c}
\end{table*}

% =====================================================================
\subsubsection{Stage 2 dataset curation prompts}
\label{ssec:s2-curation}

The Stage 2 pipeline takes each (\text{report sentence},
\text{ImaGenome phrase}, \text{anatomy}, \text{bbox}) tuple flagged as
uncertain in Stage~1 and asks the LLM to (i) rewrite the sentence so it
contains a closed-vocabulary hedge cue and an \texttt{<objN>} narrative
token, and (ii) emit a short \emph{revisit summary} describing how the
corresponding region looked when re-examined.  The output of this pass
populates the \texttt{<thinking>} portion of every Type-A Stage~2 reasoning trajectory.  \cref{tab:supp-prompt-s2} gives the prompt
template (initial call plus the self-correction continuation that fires when the verifier rejects the first output).

% ---------------------------------------------------------------------
\begin{table*}[!htbp]
\centering
\small
\renewcommand{\arraystretch}{1.25}
\begin{tabular}{@{} >{\centering\arraybackslash}p{0.13\textwidth} p{0.83\textwidth} @{}}
\toprule
\multicolumn{2}{>{\centering\arraybackslash}p{0.96\textwidth}}{\textit{(a) Initial call: hedging rewrite + visual-revisit summary (temperature $= 0.0$)}} \\
\midrule
\textbf{system content} &
You rewrite a single radiology-reasoning sentence and emit a short
visual-revisit summary.  You obey the schema strictly and never invent
findings beyond those in the provided report phrase.
\\
\midrule
\textbf{user content} &
\texttt{ORIGINAL\_SENTENCE: \{sentence\}}\newline
\texttt{IMAGENOME\_PHRASE:\hspace{2pt}\{phrase\}}\newline
\texttt{ANATOMY:\hspace{2pt}\{anatomy\}}\newline
\texttt{OBJ\_TOKEN:\hspace{2pt}\{obj\_token\}}\newline
\texttt{REVISIT\_STANCE: \{stance\}}
\par\smallskip
\textbf{Rules.}
\par
1.~\texttt{hedged\_sentence}: rewrite \texttt{ORIGINAL\_SENTENCE} so it
(a) contains \emph{at least one} hedging cue and (b) contains
\texttt{OBJ\_TOKEN} \emph{exactly once}, placed immediately after the
\texttt{ANATOMY} mention.  Preserve clinical meaning.  The hedging cue
\emph{must} be an exact substring from the closed list (case-insensitive,
no inflection---e.g.\ use ``suggests'', \emph{not} ``suggesting'' /
``suggest'' / ``suggested''; use ``appears'', \emph{not} ``appearing''):
\texttt{[\{hedges\}]}.
\par
2.~\texttt{revisit\_summary}: 1--2 sentences, references
\texttt{OBJ\_TOKEN} exactly once, with
\texttt{stance=\{stance\}} (\texttt{confirm} / \texttt{qualify} /
\texttt{alternative}).  Do \emph{not} introduce findings absent from
\texttt{IMAGENOME\_PHRASE}.  The \texttt{revisit\_summary} also
\emph{must} contain at least one exact hedging cue from the list above
---do not inflect.
\par\smallskip
\textbf{Examples} of the hedging cue used inline:\newline
\hspace*{1em}OK:~\;``The cardiac silhouette \texttt{\{obj\_token\}}
\emph{appears} enlarged \ldots''\newline
\hspace*{1em}OK:~\;``The findings in the left lung
\texttt{\{obj\_token\}} are \emph{likely} \ldots''\newline
\hspace*{1em}BAD:~``\ldots \emph{suggesting} pulmonary edema \ldots''
\;\;(inflection)\newline
\hspace*{1em}BAD:~``\ldots is \emph{suggestive} \ldots''
\;\;(must be ``suggestive of'')
\par\smallskip
Return strict JSON:
\texttt{\{"hedged\_sentence": "\ldots", "revisit\_summary": "\ldots"\}}.
\\
\midrule
\multicolumn{2}{>{\centering\arraybackslash}p{0.96\textwidth}}{\textit{(b) Self-correction turn --- fires when (a) output fails the closed-vocabulary hedge-cue verifier (temperature $= 0.4$, conversation continues from (a))}} \\
\midrule
\textbf{user content} \par\textit{\small(continuation)} &
Your previous output did not use a hedge cue from the closed list.
Re-emit the JSON.  Both fields must use one of: \emph{may, likely,
could, appears, suggests, suggestive of, concerning for, consistent
with, possible, probable, potential, perhaps}.
\\
\bottomrule
\end{tabular}
\caption{Stage 2 prompt for hedging rewrite with a self-correction fallback. 
Part (a) substitutes per-anchor variables for the original sentence, the matched ImaGenome scene-graph phrase, the anatomical region name, a unique narrative tag $\langle\text{obj}_n\rangle$, a stance label 
chosen from \{confirm, qualify, alternative\}, and a closed vocabulary of hedge terms (\textit{may}, \textit{likely}, 
\textit{could}, \textit{appears}, \textit{suggests}, \textit{suggestive of}, 
\textit{concerning for}, \textit{consistent with}, \textit{possible}, 
\textit{probable}, \textit{potential}, \textit{perhaps}). Part (b) reuses the same system content as (a). If both passes fail the 
verifier, a deterministic rule-based fallback inserts ``may'' or ``which may'' into the sentence; anchors whose rule-based output 
still fails are excluded from the SFT corpus.}
\label{tab:supp-prompt-s2}
\end{table*}

% % =====================================================================
\begin{table*}[h]
\centering
\small
\setlength{\tabcolsep}{2.5pt} 
\begin{tabular}{llcc}
\toprule
\textbf{Stage} & \textbf{Purpose} & \textbf{Input} & \textbf{Output schema} \\
\midrule
1A & Findings/Impression section parser    & full report               & \texttt{\{has\_findings|has\_impression|has\_both: bool\}} \\
1B & Sentence-level uncertainty extraction & report section            & \{\texttt{phrases, conf, type, reason}\} \\
1C & Per-phrase uncertainty classifier     & ImaGenome phrase          & \texttt{\{is\_uncertain: bool\}} \\
\addlinespace
2A & Hedging rewrite + revisit summary     & anchor record (5 fields)  & \texttt{\{hedged\_sentence, revisit\_summary\}} \\
2B & Self-correction (fallback for 2A)     & 2A output + correction    & same as 2A \\
\bottomrule
\end{tabular}
\caption{Summary of LLM calls used during data curation.  All calls target a locally hosted Qwen2.5-7B-Instruct endpoint with JSON-mode output.  Stage~1B is preceded by a regex pre-filter that already removes $\sim$60\% of obviously non-uncertain sentences, so the LLM is invoked only on borderline cases.}
\label{tab:supp-curation-summary}
\end{table*}

\subsection{Model Training Prompts}
This section lists the exact prompts used in the three training stages of UR\textsuperscript{2}-MLLM. Stage~1 is a JSON detection task; Stage~2 introduces the free-form structured-report objective with explicit reasoning; Stage~3 applies GRPO on top of Stage~2 without modifying the prompt. A
side-by-side summary is given in \cref{tab:supp-prompt-summary}.

% ---------------------------------------------------------------------
\begin{table*}[!htbp]
\centering
\small
\renewcommand{\arraystretch}{1.25}
\begin{tabular}{@{} >{\centering\arraybackslash}p{0.13\textwidth} p{0.83\textwidth} @{}}
\toprule
\textbf{system content} &
You are a radiology AI assistant.  Analyze chest X-ray images and
identify anatomical regions where the radiologist expressed diagnostic
uncertainty.
\\
\midrule
\textbf{user content} &
Analyze this chest X-ray and identify all anatomical regions with
diagnostic uncertainty.  Return a JSON list where each item contains:
\par\smallskip
\hspace*{1em}\texttt{"bbox\_2d"}:~\;\;\texttt{[x1, y1, x2, y2]}
(normalised coordinates 0--223);\newline
\hspace*{1em}\texttt{"label"}:~\;\;\;\;\;anatomical region name;\newline
\hspace*{1em}\texttt{"uncertainty"}:~the uncertain observation from the
report.
\par\smallskip
If no uncertain regions exist, return an empty list \texttt{[]}.
\\
\midrule
\textbf{assistant target} \par\textit{\small(example)} &
\texttt{[}\newline
\hspace*{1em}\texttt{\{}\newline
\hspace*{2em}\texttt{"bbox\_2d": [69, 70, 107, 108],}\newline
\hspace*{2em}\texttt{"label": "right hilar structures",}\newline
\hspace*{2em}\texttt{"uncertainty": "The hila are less prominent likely
due to \ldots"}\newline
\hspace*{1em}\texttt{\}}\newline
\texttt{]}
\\
\bottomrule
\end{tabular}
\caption{Stage 1 prompt --- uncertainty bounding-box detection (SFT).
The assistant target is a JSON array (possibly empty) of bounding-box
items.  Coordinates are normalised to the $[0, 223]$ image grid used
by the Stage~1 visual head.}
\label{tab:supp-prompt-stage1}
\end{table*}

% ---------------------------------------------------------------------
\begin{table*}[!htbp]
\centering
\small
\renewcommand{\arraystretch}{1.25}
\begin{tabular}{@{} >{\centering\arraybackslash}p{0.13\textwidth} p{0.83\textwidth} @{}}
\toprule
\textbf{system content} &
\textit{(none)}
\\
\midrule
\textbf{user content} &
\texttt{<image>}\newline
Analyze this chest X-ray and generate a structured radiology report
with explicit reasoning.  Use \texttt{<findings>}, \texttt{<thinking>},
and \texttt{<impression>} sections.
\\
\midrule
\textbf{assistant target} \par\textit{\small(example)} &
\texttt{<findings>}\newline
PA chest X-ray shows \ldots\newline
\texttt{</findings>}
\par\smallskip
\texttt{<thinking>}\newline
\ldots natural-language reasoning \ldots\newline
The hazy appearance of the right lower lung field and right hilar
lymphadenopathy \texttt{<obj1>} suggest pneumonia.\newline
\texttt{detect(}\newline
\hspace*{1em}\texttt{query="right hilar lymphadenopathy",}\newline
\hspace*{1em}\texttt{objects=["<obj1><region>[0.46,\;0.23,\;0.57,\;0.44]</region>"])}\newline
\ldots continued reasoning re.\ \texttt{<obj1>} \ldots\newline
\texttt{</thinking>}
\par\smallskip
\texttt{<impression>}\newline
Bilateral hilar lymphadenopathy and right lower lobe pneumonia \ldots\newline
\texttt{</impression>}
\\
\bottomrule
\end{tabular}
\caption{Stage 2 prompt --- reasoning-trace SFT.  The assistant target is a three-section report (\texttt{<findings>}, \texttt{<thinking>}, \texttt{<impression>}); inside \texttt{<thinking>}, uncertain regions are anchored via \texttt{<objN>} narrative tokens and inspected via
\texttt{detect()} calls whose \texttt{<region>} coordinates are
normalised to $[0, 1]$.  Type-A samples (1{,}292/1{,}909) carry at least one revisit anchor; Type-B samples (617/1{,}909) have no \texttt{detect()} calls.}
\label{tab:supp-prompt-stage2}
\end{table*}

% ---------------------------------------------------------------------
\begin{table*}[!htbp]
\centering
\small
\renewcommand{\arraystretch}{1.25}
\begin{tabular}{@{} >{\centering\arraybackslash}p{0.13\textwidth} p{0.83\textwidth} @{}}
\toprule
\textbf{system content} &
\textit{(none)}
\\
\midrule
\textbf{user content} \par\textit{\small(identical to Stage 2)} &
\texttt{<image>}\newline
Analyze this chest X-ray and generate a structured radiology report
with explicit reasoning.  Use \texttt{<findings>}, \texttt{<thinking>},
and \texttt{<impression>} sections.
\\
\midrule
\textbf{supervision} &
GRPO with the four group-relative rewards listed in this table's
caption.  No token-level target.
\\
\bottomrule
\end{tabular}
\caption{Stage 3 prompt for GRPO reinforcement learning. The user prompt is identical to Stage~2, while the token-level SFT loss is replaced by four group-relative reward functions. The \emph{format reward} 
verifies the presence and ordering of \texttt{<findings>}, 
\texttt{<thinking>}, and \texttt{<impression>}. The \emph{accuracy 
reward} measures CheXbert-style multi-label agreement with ground-truth findings on the 14 CheXpert labels. The \emph{coherence 
reward} captures consistency between \texttt{<thinking>} and the \texttt{<findings>}/\texttt{<impression>} sections. The \emph{grounding 
reward} evaluates the correctness of \texttt{<objN>}, \texttt{<region>}, and \texttt{detect()} triples and their spatial 
alignment with Stage~1 uncertainty bounding boxes.}
\label{tab:supp-prompt-stage3}
\end{table*}

% =====================================================================
\begin{table*}[!htbp]
\centering
\small
\begin{tabular}{lccc}
\toprule
 & \textbf{Stage 1} & \textbf{Stage 2} & \textbf{Stage 3} \\
\midrule
System prompt               & \checkmark             & ---                                          & --- \\
User prompt template        & detection (JSON)       & structured report                            & \textit{same as S2} \\
Output schema               & JSON array             & \texttt{<findings>/<thinking>/<impression>}  & \textit{same as S2} \\
\texttt{detect()} in target & ---                    & only Type-A samples                          & emergent under reward \\
Supervision                 & SFT (cross-entropy)    & SFT (cross-entropy)                          & GRPO (4 rewards) \\
\bottomrule
\end{tabular}
\caption{Cross-stage prompt and supervision differences. Stages~2 and 3 share an identical prompt, so all behavioral drift between them, such as the \texttt{detect()} trigger rate rising from $25.5\%$ to $74.5\%$, is attributable to reward shaping rather than prompt engineering.}
\label{tab:supp-prompt-summary}
\end{table*}

\section{Uncertainty Analysis}

\begin{figure*}
    \centering 
    \includegraphics[width=1\textwidth]{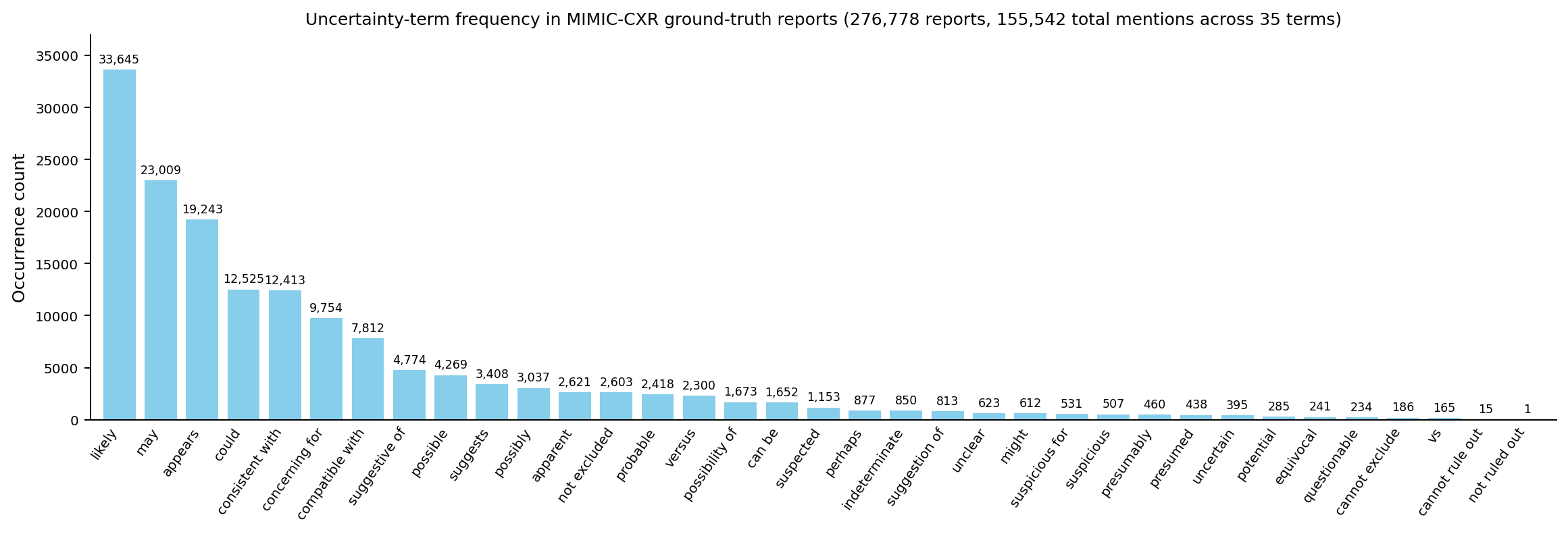} 
    % \caption{Overall framework of UR\textsuperscript{2}-MLLM, including (a) data curation pipeline, (b) training stage overview, and (c) the three-stage training procedure.}  
    \caption{Frequency distribution of uncertainty markers in MIMIC-CXR ground-truth reports.  All $276{,}778$ reports (train+val+test) were scanned for occurrences of the $35$-term closed vocabulary used during Stage~2 trace synthesis (longest-first masked matching, case-insensitive, whole-word boundary).  Top three markers (\textit{likely}, \textit{may}, \textit{appears}) cover $48.8\%$ of all $155{,}542$ mentions; the top ten cover $84\%$.  Multi-word constructions (\textit{consistent with}, \textit{concerning for}, \textit{compatible with}) appear prominently among the top seven.}
    \label{fig:uncertainty}  
\end{figure*}

\cref{fig:uncertainty} reports the frequency distribution of uncertainty markers in the full MIMIC-CXR corpus, which contains $276{,}778$ reports and $13.3$\,M whitespace tokens.  We scan every report for occurrences of a closed vocabulary of 35 predefined hedge terms, using case-insensitive whole-word matching with longest-first masking, so that multi-word terms such as \textit{suggestive of} are not double-counted by their constituent single words. In total we observe $155{,}542$ hedge instances, corresponding to $0.56$ mentions per report on average, or $1.17\%$ of all word tokens.  Bars are sorted in descending order of count. The distribution is heavily long-tailed.  The three most frequent markers, namely \textit{likely} ($33{,}645$), \textit{may} ($23{,}009$), and \textit{appears} ($19{,}243$), account for $48.8\%$ of all mentions, while the top ten cover $84\%$ and the bottom ten contribute fewer than $3{,}000$ mentions combined.  Multi-word constructions such as \textit{consistent with}, \textit{concerning for}, \textit{compatible with}, and \textit{suggestive of} appear prominently among the top eight, together accounting for $22.3\%$ of all hedge mentions.  All $35$ terms in the vocabulary appear at least once.

\begin{figure*}
    \centering 
    \includegraphics[width=1\textwidth]{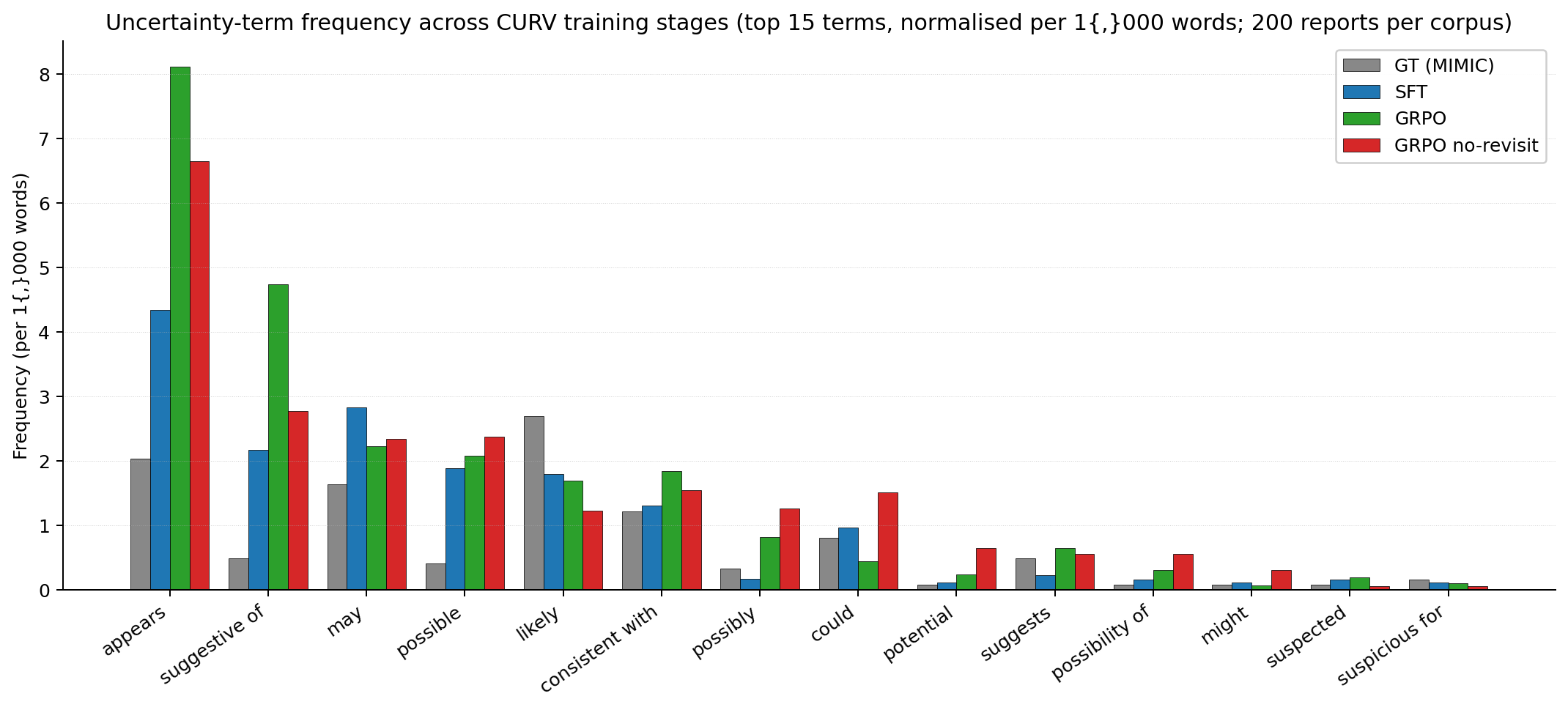} 
    % \caption{Overall framework of UR\textsuperscript{2}-MLLM, including (a) data curation pipeline, (b) training stage overview, and (c) the three-stage training procedure.}  
    \caption{Per-thousand-word frequency of the top fourteen uncertainty markers across four corpora.  Bars within each group correspond, from left to right, to ground-truth MIMIC-CXR reports, the Stage 2 SFT model, the full-reward GRPO model, and the no-revisit GRPO ablation.  All four corpora are matched on the same 200-report test split, with rates normalised by total word count.}
    \label{fig:top15}  
\end{figure*}

\cref{fig:top15} reports per-thousand-word frequencies for the fourteen most common uncertainty markers across four corpora, namely the MIMIC-CXR ground-truth reports (GT), the Stage~2 SFT model, the full-reward GRPO model, and the no-revisit GRPO ablation. Comparing these distributions reveals how the hedging vocabulary narrows and reshapes as training proceeds.

The ground-truth corpus exhibits a relatively balanced spread of
hedging cues.  Its five most common markers are \emph{likely}
(2.69 per thousand words), \emph{appears} (2.03), \emph{may} (1.63), \emph{consistent with} (1.22), and \emph{could} (0.81), where the most frequent term is only about thirty times more common than the least frequent one shown and no single cue dominates radiologist phrasing.

This balance does not survive training.  All three model variants shift the modal hedge from \emph{likely}, which leads in the GT, to \emph{appears}, which becomes the single most frequent marker in every model corpus, and inflate the rates of a small set of generic cues.  Already at the SFT stage, the model uses \emph{appears} at more than twice the GT rate (4.34 against 2.03), \emph{suggestive of} at more than four times the GT rate (2.17 against 0.49), and \emph{possible} at a comparable margin (1.89 against 0.41).  The full-reward GRPO model further amplifies the same cues, with \emph{appears} rising to 8.11 and \emph{suggestive of} to 4.74, while
the no-revisit ablation reproduces the same pattern at a moderated magnitude (6.65, 2.77, and 2.37 for the three terms). This indicates that the over-reliance on a small group of generic hedges is established during Stage~2 SFT and then further accentuated by the GRPO objective rather than introduced by it.

The revisit reward affects this picture not at the top of the
distribution but in its mid-frequency range.  For \emph{could},
\emph{possibly}, \emph{possibility of}, \emph{might}, and
\emph{potential}, the no-revisit model produces noticeably higher rates than the full-reward GRPO, and in several cases exceeds the GT rate as well, with \emph{could} at 1.51 against 0.81 in the GT, \emph{possibly} at 1.26 against 0.33, and \emph{might} at 0.31 against 0.08.  In contrast, the full-reward GRPO concentrates its hedging on the most frequent markers and uses these mid-tail cues sparingly.  Removing the revisit reward therefore broadens the hedging vocabulary while leaving the top-end over-use largely intact.

\section{Reasoning Evaluation}
\subsection{LLM-based Evaluation Prompts}

We score each generated trace on five dimensions
(\cref{tab:supp-prompt-llm-judge}) using GPT-4o as judge.
Each dimension receives an integer score from $1$ to $5$ together with a one-sentence justification; aggregate means are reported in \cref{tab:thinking-eval-scores}.

\subsection{LLM-based Reasoning Evaluation}

\cref{tab:thinking-eval-scores} reports the GPT-as-judge scores for the \texttt{<thinking>} portion of generated reports, averaged over $50$ randomly chosen test samples per model.  The three model variants differ mainly in which dimension they favor rather than in overall quality.  The Stage~2 SFT model achieves the highest clinical plausibility ($2.26$ against $2.02$ for GRPO and $1.92$ for no-revisit) and the highest faithfulness ($2.38$ against $2.26$ and
$2.20$), indicating that the SFT objective keeps the model closest to the hedged phrasing of the curated training traces.  The full-reward GRPO model instead favors uncertainty calibration, scoring slightly higher than the other variants ($2.56$ against $2.38$ for SFT and $2.50$ for no-revisit) and echoing the calibration improvement observed in the per-finding analysis above.  The no-revisit ablation scores highest on coherence ($3.20$) and on the findings-to-impression bridge ($2.18$). Both advantages are partly explained by its longer outputs, a known length bias in LLM-as-judge evaluation.  Since all three models cluster within $0.07$ of each other on the AVERAGE row, well inside one pooled standard error, the table is best read as a per-dimension diagnostic rather than a global ranking.  Taken together, the GRPO objective appears to trade a small amount of faithfulness for better-calibrated hedging, consistent with the per-finding calibration analysis.

\section{Case Study}
\label{app:casestudy}

\begin{figure*}[t]
    \centering
    \begin{subfigure}[b]{0.95\textwidth}
        \centering
        \includegraphics[width=\textwidth]{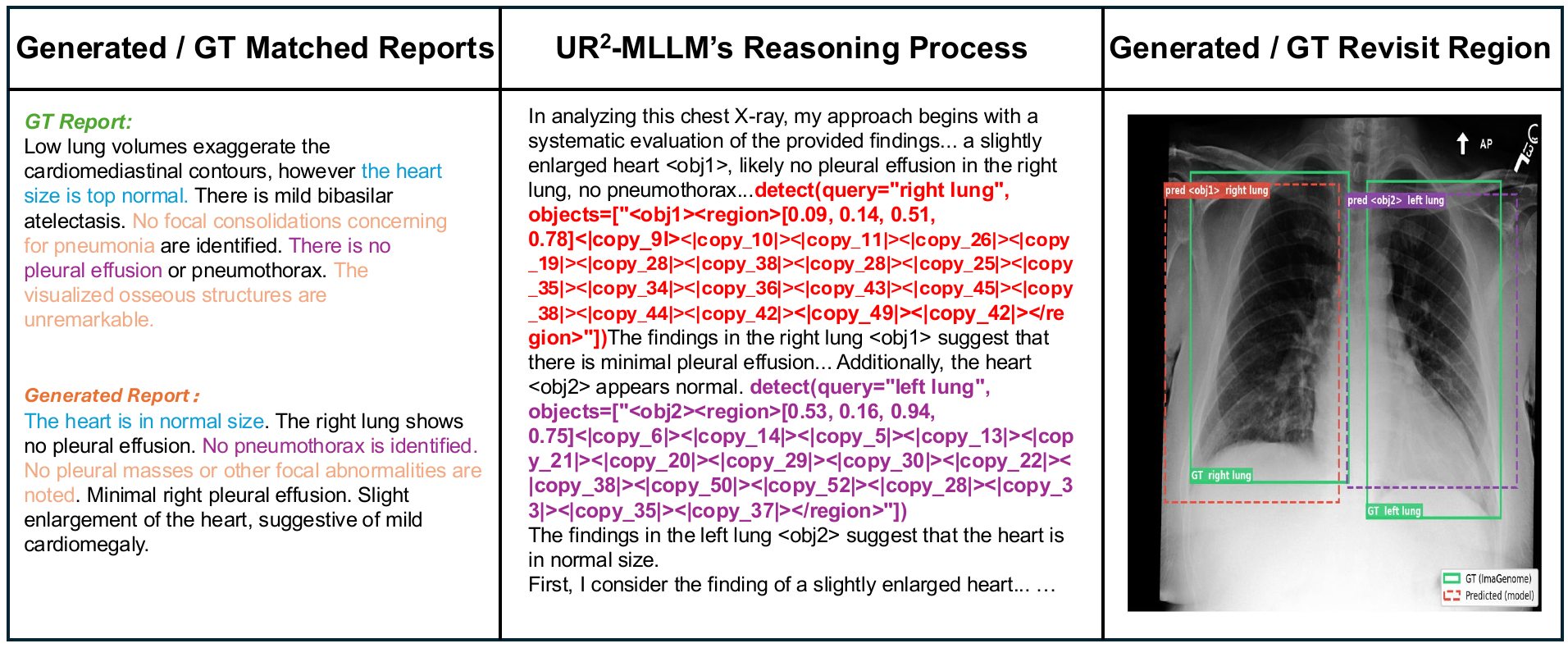}
        \caption{Success sample case.}
        \label{fig:main_a}
    \end{subfigure}
    
    \begin{subfigure}[b]{0.95\textwidth}
        \centering
        \includegraphics[width=\textwidth]{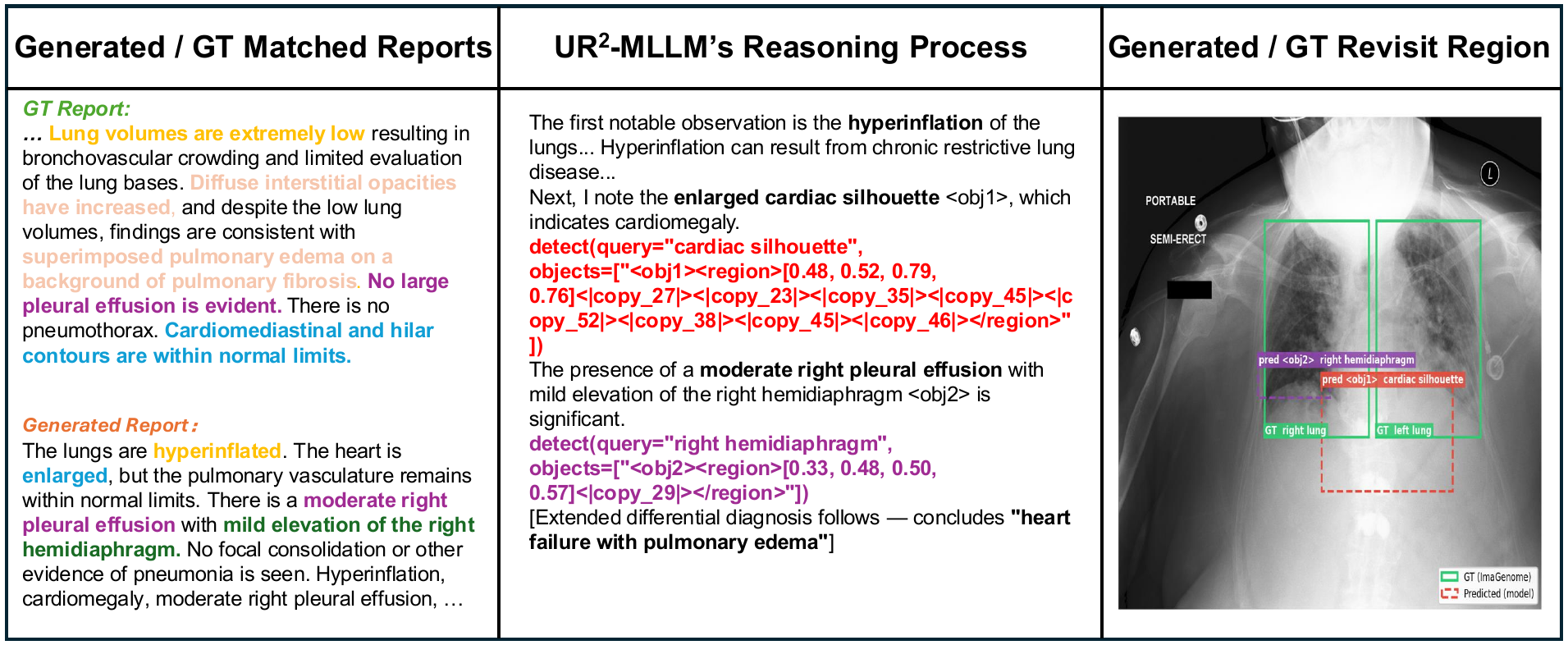}
        \caption{Failure sample case.}
        \label{fig:main_b}
    \end{subfigure}
    
    \caption{Two qualitative case studies from the MIMIC-CXR.}
    \label{fig:case-study}
\end{figure*}

To better understand the behavior of UR\textsuperscript{2}-MLLM, we present two qualitative case studies from MIMIC-CXR in \cref{fig:case-study}, illustrating both a successful prediction and a more challenging failure scenario. 

\cref{fig:case-study}(a) shows a successful case where the model correctly handles a clinically subtle finding. The ground-truth report notes that low lung volumes exaggerate the cardiomediastinal contours while the heart size remains top normal. Our model reaches the same conclusion through an explicit reasoning trace. It first detects the right lung and the left lung as separate regions, grounds them via bounding box predictions, and then evaluates the cardiac silhouette within the context of the surrounding lung fields. By revisiting these anatomical regions before issuing a final judgment, the model avoids the common pitfall of over-pathologizing a low-volume study and produces a generated report that closely aligns with the ground truth on heart size, pneumothorax, and pleural effusion. The predicted bounding boxes also overlap well with the ImaGenome ground-truth regions, indicating that the visual grounding is consistent with the textual reasoning. 

\cref{fig:case-study}(b) presents a more challenging case in which the underlying pathology is severe and involves multiple coexisting findings. The ground-truth report describes extremely low lung volumes, diffuse interstitial opacities, and superimposed pulmonary edema on a background of pulmonary fibrosis, with cardiomediastinal and hilar contours remaining within normal limits. Our model correctly identifies the hyperinflation pattern and grounds both the cardiac silhouette and the right hemidiaphragm through its reasoning process, yet it ultimately commits to a differential diagnosis of heart failure with pulmonary edema and reports cardiomegaly together with a moderate right pleural effusion that the radiologist does not confirm. Although the final output is incorrect, the reasoning trace itself remains valuable. It reveals exactly where the error originates, namely at the transition from regional findings to the global differential diagnosis, where the model over-commits to a familiar disease pattern instead of staying faithful to the visually grounded evidence. This kind of transparency is difficult to obtain from end-to-end models and points to a clear direction for future improvement, which is to introduce stronger consistency constraints between the grounded regions and the final diagnostic conclusion.

\begin{table*}[!htbp]
\centering
\small
\renewcommand{\arraystretch}{1.25}
\begin{tabular}{@{} >{\centering\arraybackslash}p{0.18\textwidth} p{0.78\textwidth} @{}}
\toprule
\textbf{system content} &
You are a senior radiologist evaluating an AI-generated chest X-ray reasoning trace.  You are given the ground-truth
(radiologist-written) report and an AI-generated output that contains
\texttt{<findings>}, \texttt{<thinking>}, and \texttt{<impression>}
sections.  You will score the AI output on five dimensions on a
$1$--$5$ 5-point rating scale.  Output strict JSON only.  Length and writing style are \emph{not} criteria.  You do not know which AI model produced the output.
\\
\midrule
\multicolumn{2}{>{\centering\arraybackslash}p{0.96\textwidth}}{\textit{User content (scoring rubric, $1$--$5$ integer per dimension; higher is better)}} \\
\midrule
\textbf{1. clinical plausibility} &
How medically sensible is the reasoning given a chest X-ray?\newline
\hspace*{1em}1 = medically incorrect or nonsensical.\newline
\hspace*{1em}3 = mostly plausible with notable inaccuracies.\newline
\hspace*{1em}5 = fully clinically valid; a radiologist would accept
the inferences.
\\
\midrule
\textbf{2. faithfulness} &
Compare claims in the AI output (especially in \texttt{<findings>}
and \texttt{<thinking>}) against the ground-truth report.\newline
\hspace*{1em}1 = severely hallucinated; the trace asserts findings
clearly absent from the GT.\newline
\hspace*{1em}3 = partially grounded; some unsupported claims.\newline
\hspace*{1em}5 = fully grounded; no findings or locations introduced
beyond what the GT supports.
\\
\midrule
\textbf{3. coherence} &
Internal logical consistency of \texttt{<thinking>}.\newline
\hspace*{1em}1 = self-contradictory, jumps between unrelated
claims.\newline
\hspace*{1em}3 = mostly coherent, with some logical gaps.\newline
\hspace*{1em}5 = clean logical progression from observation to
inference, no contradictions.
\\
\midrule
\textbf{4. findings$\rightarrow$impression bridge} &
How well does \texttt{<thinking>} motivate \texttt{<impression>}?\newline
\hspace*{1em}1 = impression conclusions are not supported by
\texttt{<thinking>}.\newline
\hspace*{1em}3 = some impression statements are supported, others are
not.\newline
\hspace*{1em}5 = every claim in \texttt{<impression>} is traceable to
reasoning in \texttt{<thinking>}.
\\
\midrule
\textbf{5. uncertainty calibration} &
Are hedge words (\textit{likely, may, appears, suggestive of,
possible, consistent with}, \ldots) and confident assertions placed
appropriately?\newline
\hspace*{1em}1 = hedges used randomly or missing where evidence is
weak.\newline
\hspace*{1em}3 = some hedges are warranted, others misplaced.\newline
\hspace*{1em}5 = hedges only where evidence is genuinely uncertain;
commitment where evidence is strong.
\\
\midrule
\multicolumn{2}{>{\centering\arraybackslash}p{0.96\textwidth}}{\textit{User content (input substitution and required output)}} \\
\midrule
\textbf{inputs} &
\texttt{=== GROUND-TRUTH REPORT (written by a radiologist) ===}\newline
\texttt{\{gt\_report\}}\newline
\newline
\texttt{=== AI-GENERATED OUTPUT (contains <findings>, <thinking>, <impression>) ===}\newline
\texttt{\{generated\}}
\\
\midrule
\textbf{output schema} &
Return a single JSON object (no markdown, no trailing text):\newline
\texttt{\{}\newline
\hspace*{1em}\texttt{"clinical\_plausibility":\hspace{3pt}\{"score": <1-5>, "justification": "\ldots"\},}\newline
\hspace*{1em}\texttt{"faithfulness":\hspace{12pt}\{"score": <1-5>, "justification": "\ldots"\},}\newline
\hspace*{1em}\texttt{"coherence":\hspace{20pt}\{"score": <1-5>, "justification": "\ldots"\},}\newline
\hspace*{1em}\texttt{"bridge":\hspace{30pt}\{"score": <1-5>, "justification": "\ldots"\},}\newline
\hspace*{1em}\texttt{"uncertainty\_calibration":\hspace{2pt}\{"score": <1-5>, "justification": "\ldots"\}}\newline
\texttt{\}}
\\
\bottomrule
\end{tabular}
\caption{LLM-as-judge prompt used to score the \texttt{<thinking>} portion of generated reasoning traces.  The judge is GPT-4o (\texttt{gpt-4o-2024-11-20}) called with $\text{temperature}=0$ and JSON-mode output; identical \texttt{seed} is used across all reports to ensure reproducibility.  For every test report and every model under evaluation, the prompt is instantiated by substituting the radiologist-written ground-truth report into \texttt{\{gt\_report\}} and the model's full \texttt{<findings>/<thinking>/<impression>}
generation into \texttt{\{generated\}}.  All five dimensions are scored on a $1$--$5$ Likert scale; each score is accompanied by a one-sentence justification. Aggregate results are reported in \cref{tab:thinking-eval-scores}.}
\label{tab:supp-prompt-llm-judge}
\end{table*}

\begin{table*}[!htbp]
\centering
\small
\setlength{\tabcolsep}{3pt}
\renewcommand{\arraystretch}{1.25}
\begin{tabular}{@{} l c c c @{}}
\toprule
\textbf{Dimension}                        & \textbf{SFT}                  & \textbf{GRPO}                 & \textbf{GRPO no-revisit}      \\
\midrule
Clinical plausibility                     & \textbf{$2.26\pm0.60$}        & $2.02\pm0.59$                 & $1.92\pm0.67$                 \\
Faithfulness                              & \textbf{$2.38\pm0.53$}        & $2.26\pm0.44$                 & $2.20\pm0.45$                 \\
Coherence                                 & $3.06\pm0.74$                 & $3.00\pm0.53$                 & \textbf{$3.20\pm0.40$}        \\
Findings$\rightarrow$Impression bridge    & $2.16\pm0.74$                 & $2.06\pm0.55$                 & \textbf{$2.18\pm0.52$}        \\
Uncertainty calibration                   & $2.38\pm0.67$                 & \textbf{$2.56\pm0.64$}        & $2.50\pm0.65$                 \\
\midrule
\textbf{Average}                          & \textbf{$2.45$}               & $2.38$                        & $2.40$                        \\
\bottomrule
\end{tabular}
\caption{Mean $\pm$ std of GPT-4o-as-judge scores ($1$ to $5$) on the \texttt{<thinking>} portion ($n=50$ per model).  ``Bridge'' denotes how well \texttt{<thinking>} supports
\texttt{<impression>}.  Best in bold; SE of the mean $\leq 0.11$. Full prompt in \cref{tab:supp-prompt-llm-judge}.}
\label{tab:thinking-eval-scores}
\end{table*}

\section{Licenses for External Assets}
\label{app:licenses}
This appendix summarizes the sources, URLs, licenses, and access conditions of the publicly available datasets and pre-trained models used in this work. Details can be found in \cref{tab:licenses}.

\begin{table*}[t]
\centering
\small
\setlength{\tabcolsep}{10pt}
\renewcommand{\arraystretch}{1.15}
\begin{tabularx}{0.96\textwidth}{|X|}
\hline

\textbf{Datasets} \\
\hline

\begin{itemize}[leftmargin=2.0em, itemsep=0.6em]

    \item \textbf{MIMIC-CXR (version 2.1.0):}
    \begin{itemize}[leftmargin=2.0em, itemsep=0.15em]
        \item[\textbf{--}] \textbf{Source:} Johnson~\etal~\cite{PhysioNet-mimic-cxr-2.1.0}.
        \item[\textbf{--}] \textbf{URL:} \url{https://physionet.org/content/mimic-cxr/2.1.0/}
        \item[\textbf{--}] \textbf{License:} PhysioNet Credentialed Health Data License 1.5.0. Access requires credentialing and signing a data use agreement.
    \end{itemize}

    \item \textbf{Chest ImaGenome Dataset (version 1.0.0):}
    \begin{itemize}[leftmargin=2.0em, itemsep=0.15em]
        \item[\textbf{--}] \textbf{Source:} Wu~\etal~\cite{PhysioNet-chest-imagenome-1.0.0}.
        \item[\textbf{--}] \textbf{URL:} \url{https://physionet.org/content/chest-imagenome/1.0.0/}
        \item[\textbf{--}] \textbf{License:} Derived from MIMIC-CXR, subject to the PhysioNet Credentialed Health Data License 1.5.0.
    \end{itemize}

\end{itemize}
\\
\hline

\textbf{Pre-trained Models and Baselines} \\
\hline

\begin{itemize}[leftmargin=2.0em, itemsep=0.6em]

    \item \textbf{Qwen-2.5-VL-3B (Backbone for UR\textsuperscript{2}-MLLM):}
    \begin{itemize}[leftmargin=2.0em, itemsep=0.15em]
        \item[\textbf{--}] \textbf{Source:} Bai~\etal~\cite{DBLP:journals/corr/abs-2511-21631}.
        \item[\textbf{--}] \textbf{URL:} \url{https://github.com/QwenLM/Qwen3}
        \item[\textbf{--}] \textbf{License:} Apache 2.0 License.
    \end{itemize}

    \item \textbf{LLaVA-1.5-7B:}
    \begin{itemize}[leftmargin=2.0em, itemsep=0.15em]
        \item[\textbf{--}] \textbf{Source:} Liu~\etal~\cite{liu2024improved}.
        \item[\textbf{--}] \textbf{URL:} \url{https://github.com/haotian-liu/LLaVA}
        \item[\textbf{--}] \textbf{License:} Apache 2.0 License.
    \end{itemize}

    \item \textbf{MAIRA-2:}
    \begin{itemize}[leftmargin=2.0em, itemsep=0.15em]
        \item[\textbf{--}] \textbf{Source:} Bannur~\etal~\cite{DBLP:journals/corr/abs-2406-04449}.
        \item[\textbf{--}] \textbf{URL:} \url{https://huggingface.co/microsoft/maira-2}
        \item[\textbf{--}] \textbf{License:} Microsoft Research License Terms.
    \end{itemize}

    \item \textbf{HuatuoGPT-Vision-7B:}
    \begin{itemize}[leftmargin=2.0em, itemsep=0.15em]
        \item[\textbf{--}] \textbf{Source:} Chen~\etal~\cite{chen-etal-2024-towards-injecting}.
        \item[\textbf{--}] \textbf{URL:} \url{https://huggingface.co/FreedomIntelligence/HuatuoGPT-Vision-7B}
        \item[\textbf{--}] \textbf{License:} Apache 2.0 License.
    \end{itemize}

    \item \textbf{RadGraph:}
    \begin{itemize}[leftmargin=2.0em, itemsep=0.15em]
        \item[\textbf{--}] \textbf{Source:} Jain~\etal~\cite{PhysioNet-radgraph-1.0.0}.
        \item[\textbf{--}] \textbf{URL:} \url{https://huggingface.co/StanfordAIMI/RRG_scorers}
        \item[\textbf{--}] \textbf{License:} MIT License.
    \end{itemize}

    \item \textbf{CheXbert:}
    \begin{itemize}[leftmargin=2.0em, itemsep=0.15em]
        \item[\textbf{--}] \textbf{Source:} Smit~\etal~\cite{smit2020combining}.
        \item[\textbf{--}] \textbf{URL:} \url{https://huggingface.co/StanfordAIMI/RRG_scorers}
        \item[\textbf{--}] \textbf{License:} MIT License.
    \end{itemize}

\end{itemize}
\\
\hline

\end{tabularx}
\caption{Licenses and access conditions of the publicly available datasets and pre-trained models used in this work.}
\label{tab:licenses}
\end{table*}

\end{document}